\documentclass[11pt]{article}

\usepackage[12pt]{extsizes}
\usepackage[utf8]{inputenc}
\usepackage{amsmath}
\usepackage{amssymb}
\usepackage{amsthm}
\usepackage{amsfonts}
\usepackage[numbers,sort&compress]{natbib}
\usepackage{graphicx}
\usepackage{lipsum}
\usepackage{enumitem}
\usepackage[font=footnotesize,labelfont=bf]{caption}
\usepackage{mdframed}
\usepackage{authblk}
\usepackage{longtable}
\usepackage{subfigure}
\usepackage[subfigure]{tocloft}
\usepackage[dvipsnames,table]{xcolor}
\usepackage[hidelinks]{hyperref}
\usepackage{mathtools}
\usepackage{relsize}
\usepackage{multirow}
\usepackage{booktabs}
\usepackage{tabularx}
\usepackage{bbding}
\usepackage{listings}
\usepackage{xspace}
\usepackage[ruled,vlined]{algorithm2e}
\usepackage{multirow}
\usepackage[p,osf]{cochineal}
\usepackage[scale=.95,type1]{cabin}
\usepackage[cochineal,bigdelims,cmintegrals,vvarbb]{newtxmath}
\usepackage[zerostyle=c,scaled=.94]{newtxtt}
\usepackage[cal=boondoxo]{mathalfa}
\usepackage{microtype}
\usepackage{multirow}
\usepackage{adjustbox}
\usepackage{etoolbox}
\usepackage{csquotes}
\usepackage[title]{appendix}%

\usepackage{caption}
\usepackage{geometry}
\definecolor{myurlcolor}{HTML}{123463}
\definecolor{dc_color}{RGB}{230, 245, 244}
\definecolor{ds_color}{RGB}{195, 230, 227}
\definecolor{ms_color}{RGB}{150, 214, 209}
\hypersetup{
    colorlinks=true,
    linkcolor=red,
    filecolor=black,      
    urlcolor=black,
    citecolor=blue
}

\apptocmd{\thebibliography}{\raggedright}{}{}

\makeatletter
\patchcmd{\@maketitle}{\LARGE \@title}{\fontsize{30}{19.2}\selectfont\@title}{}{}
\makeatother
\usepackage{cleveref}
\Crefname{section}{Sec.}{Secs.}
\Crefname{equation}{Eq.}{Eqs.}
\Crefname{figure}{Figure}{Figs.}
\Crefname{tabular}{Table}{Tabs.}

\usepackage{tabularx}
\usepackage{ragged2e}
\usepackage{lineno}
\usepackage{chngcntr}
\usepackage{tcolorbox}

\PassOptionsToPackage{table,x11names}{xcolor}
\usepackage{pifont}
\newcommand{\cmark}{\textcolor{green!60!black}{\ding{51}}}  
\newcommand{\xmark}{\textcolor{red}{\ding{55}}}             

\newif\ifhighlighton

\highlightonfalse 

\ifhighlighton
    \definecolor{customdefaultcolor}{RGB}{18, 159, 87}
    \NewDocumentCommand{\highlight}{O{customdefaultcolor} +m}{%
        \begingroup
        \color{#1}#2%
        \endgroup
    }
\else
    \NewDocumentCommand{\highlight}{O{customdefaultcolor} +m}{#2}
\fi

\usepackage{titletoc}
\newcommand\DoToC{%
  \startcontents
  \printcontents{}{1}{\textbf{Supplementary Notes}\vskip3pt\hrule\vskip5pt}
  \vskip3pt\hrule\vskip5pt
}

\usepackage{minted}
\setminted{
  breaklines,
  breakanywhere,   
  autogobble,      
  fontsize=\scriptsize,
  tabsize=2
}

\usepackage{listings}
\definecolor{bggray}{rgb}{0.97,0.97,0.97}
\definecolor{bordergray}{rgb}{0.85,0.85,0.85}
\definecolor{textgray}{rgb}{0.25,0.25,0.25}
\definecolor{keywordcolor}{rgb}{0.35,0.45,0.85}
\definecolor{commentcolor}{rgb}{0.2,0.6,0.2}
\definecolor{stringcolor}{rgb}{0.65,0.1,0.1}

\usepackage{setspace}

\begin{document}

\newcommand{\datasetname}{\texttt{TrialPanorama}\xspace}

\title{\LARGE \textbf{Developing Large Language Models for Clinical Research Using One Million Clinical Trials}}

\author[1+]{Zifeng Wang}
\author[2+]{Jiacheng Lin}
\author[3+]{Qiao Jin}
\author[4,5]{Junyi Gao}
\author[2]{Jathurshan Pradeepkumar}
\author[2]{Pengcheng Jiang}
\author[3]{Zhiyong Lu}
\author[1,2\#]{Jimeng Sun}

\affil[1]{Keiji AI, Seattle, USA}
\affil[2]{School of Computing and Data Science, University of Illinois Urbana-Champaign, Urbana, IL, USA}
\affil[3]{Division of Intramural Research, National Library of Medicine, National Institutes of Health, Bethesda, MD, USA}
\affil[4]{Usher Institute, Edinburgh Medical School, University of Edinburgh, Edinburgh, UK}
\affil[5]{Health Data Research UK, UK}

\affil[+]{\em{Equal contribution}}
\affil[$\#$]{\em{Correspondence: \href{mailto:zifeng@keiji.ai}{zifeng@keiji.ai}, \href{mailto:jimeng@illinois.edu}{jimeng@illinois.edu}}}


\date{\vspace{-0.5em}\small\itshape
Project website: \url{https://keiji.ai/research/trialpanorama}
}

\maketitle



\abstract{
Developing artificial intelligence (AI) for clinical research requires a comprehensive data foundation that supports model training and rigorous evaluation. Here, we introduce \datasetname, a large-scale structured resource that aggregates 1.6M clinical trial records from fifteen global registries and links them with biomedical ontologies and associated literature. To demonstrate its utility, we build a pipeline that constructs 152K training and testing samples for eight key clinical research tasks. Three tasks support systematic review workflows, including study search, study screening, and evidence summarization. Five tasks focus on trial design and optimization, including arm design, eligibility criteria design, endpoint selection, sample size estimation, and trial completion assessment and rationalization. Benchmarking cutting-edge large language models (LLMs) reveals that generic LLMs have limited capability in clinical reasoning. In contrast, an 8B LLM we developed on \datasetname using supervised finetuning and reinforcement learning wins over the 70B generic counterparts in all eight tasks, with a relative improvement of 73.7\%, 67.6\%, 38.4\%, 37.8\%, 26.5\%, 20.7\%, 20.0\%, 18.1\%, and 5.2\%, respectively. We envision \datasetname provides a solid foundation for future scaling of AI for clinical research.

}


\date{}

\maketitle




\section*{Introduction}
Developing new drugs remains a lengthy and expensive process that often spans close to a decade and costs more than two billion dollars~\cite{martin2017much}. A major share of this burden arises during the human study phase when clinical trials are conducted to evaluate safety and efficacy across phases I through IV. Beyond the execution of trials, clinicians, researchers, and trial designers routinely search for and review clinical trial protocols and results. These activities support essential objectives such as planning new studies, assessing treatments, and informing clinical decision making in evidence-based medicine~\cite{field2010meta,concato2017randomized}.

There is growing interest in using artificial intelligence (AI), and especially large language models (LLMs), to improve the development and evaluation of clinical trials. Recent advances have explored a diverse set of tasks including predicting clinical trial outcomes~\cite{jin2020predicting,fu2022hint,wang2023spot}, matching patients to appropriate trials~\cite{jin2024matching,wornow2025zero}, extracting real world cohort data from eligibility criteria~\cite{yuan2019criteria2query,park2024criteria2query}, drafting clinical trial documents~\cite{wang2023autotrial,wang2025informgen}, and accelerating systematic review of clinical studies~\cite{wang2025foundation,wang2024accelerating}. Nonetheless, most depend on specialized data sources assembled from various specialized repositories of clinical trials. As a result, progress is constrained by the absence of standardized and comprehensive clinical trial resources suitable for AI development.

In this work, we introduce \datasetname, a clinical trial database and benchmark suite created from over one million records and fifteen global sources, to provide a broad foundation for trial design, trial review, and the development of AI for clinical trial tasks. We assemble clinical trial records from major global registries, including ClinicalTrials.gov~\cite{clinicaltrialsgov}, the International Clinical Trials Registry Platform~\cite{whoictrp}, the European Clinical Trials Register~\cite{euclinicaltrial}, and several national registries~\cite{anzctr,chinesechictr}. We further incorporate trial-related evidence from literature sources such as PubMed. Prior studies have curated selected portions of these resources for focused objectives such as extracting trial results~\cite{shi2024constructing,naik2024care}, annotating drugs and conditions~\cite{chen2022knowledge,brbic2024predicting}, linking publications to registered trials~\cite{xu2020building}, and generating structured outcome data~\cite{gao2024automatically}. Here, we put them into a unified representation that integrates these diverse elements, which can substantially strengthen evidence synthesis in systematic reviews and support data driven approaches to future trial planning~\cite{dunn2020time,holzhauer2023super,kim2017broadening}.

Specifically, \datasetname is created as both a database and a tool suite for extracting clinical trial information and producing benchmark data for AI development in clinical trial applications. To illustrate its utility, we present a pipeline that extracts training and testing data from the resource and supports eight trial-related tasks. Five tasks focus on trial design and optimization, including arm design, eligibility criteria design, endpoint design, sample size estimation, and trial completion reasoning and prediction. Three tasks support systematic literature review workflows, including study search, study screening, and evidence question answering. We further evaluate open source and proprietary LLMs on these tasks to establish baseline performance and to demonstrate the value of \datasetname for future research.

Our benchmark further demonstrates the limitations of generic LLMs when applied to clinical trial tasks. We find that widely used LLMs display substantial gaps in study search, study screening, evidence summarization, and several trial design and optimization tasks. To explore the value of domain-specific data, we fine-tune two open-source models, Llama-3 ~\cite{dubey2024llama} and Qwen-3 ~\cite{yang2025qwen3}, using training data extracted from \datasetname. These tuned models achieve competitive or superior performance relative to larger generic LLMs across both systematic review and trial design tasks. This improvement highlights the importance of high-quality trial-specific data foundations and demonstrates that \datasetname can meaningfully strengthen the capabilities of LLMs developed for clinical trial applications.
\section*{Results}

\subsection*{\datasetname: a unified database and knowledge graph of clinical trials}
\begin{figure}
    \centering
    \includegraphics[width=0.95\linewidth]{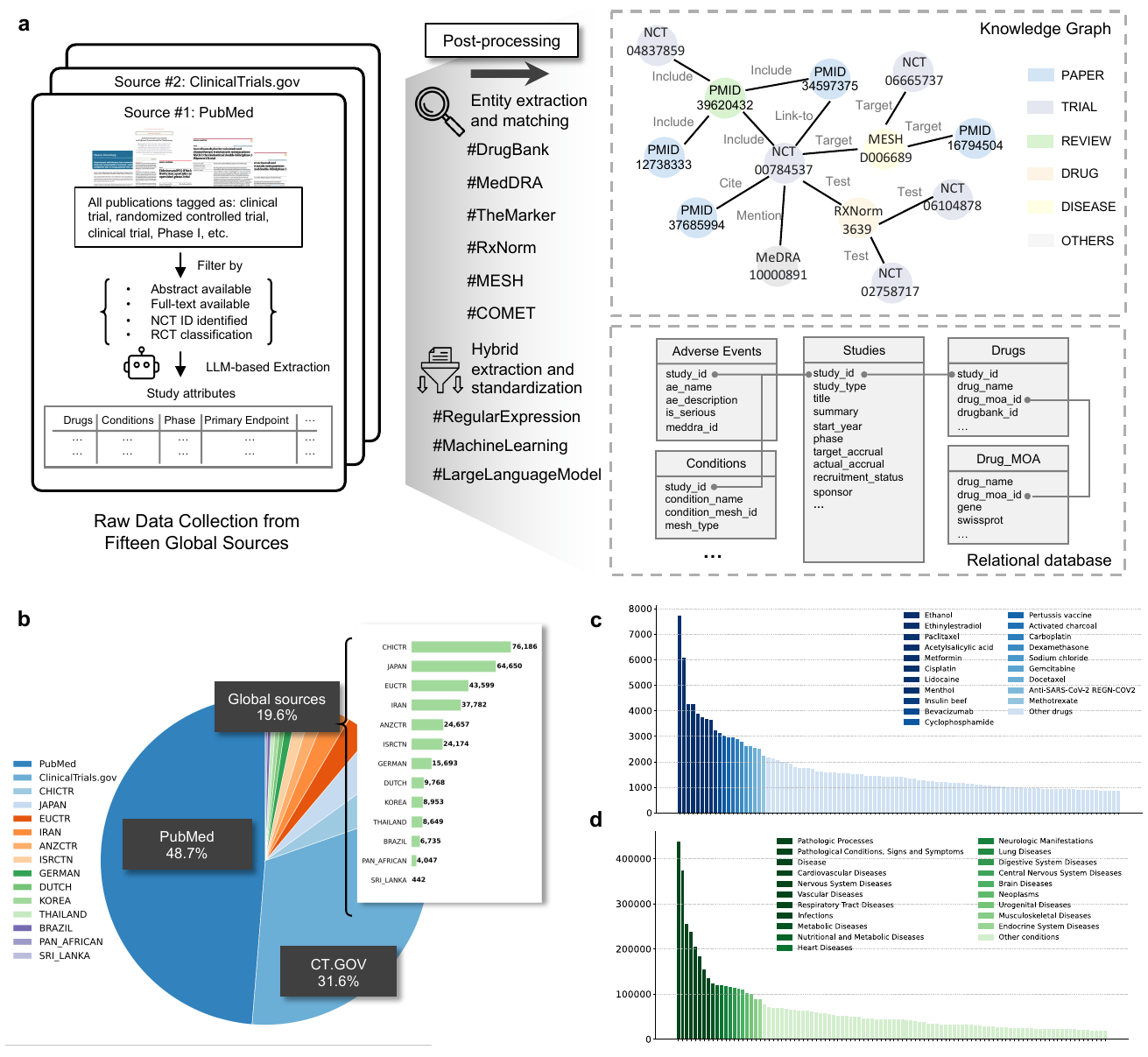}
    \caption{Overview of \datasetname and its end-to-end processing workflow.
\textbf{a}, Raw clinical research data are collected from fifteen global sources, including PubMed, ClinicalTrials.gov, and international registries. LLM-based extraction, hybrid rule-based methods, and ontology-grounded normalization are applied to derive unified study attributes, followed by entity extraction and matching to DrugBank, MedDRA, RxNorm, MeSH, and other biomedical ontologies. The processed information is stored in both a relational schema and a heterogeneous knowledge graph linking trials, publications, drugs, diseases, and outcomes. \textbf{b}, Distribution of data sources in \datasetname showing contributions from PubMed, ClinicalTrials.gov, and aggregated global registries, together with a country-level breakdown of non-US registries. \textbf{c}, Most frequent drug entities mapped to standardized identifiers, illustrating the long tail distribution and coverage of therapeutic compounds. \textbf{d}, Most frequent disease and condition entities after ontology harmonization, demonstrating broad coverage across oncology, infectious disease, neurologic disease, metabolic disorders, cardiovascular disease, and other major therapeutic areas.}
    \label{fig:figure1}
\end{figure}

Public information on clinical trials is available mainly through research publications and registry sites. For publications, we collected records from PubMed by focusing on papers labeled as randomized controlled trials and related tags, then applied a machine learning classifier to identify those that truly report clinical trial studies. For registry sources, we gathered data from major countries and regions worldwide, including the large registry clinicaltrials.gov and others such as CHICTR in China. Together, these sources yield more than 1.6 million trial records. As shown in Figure~\hyperref[fig:figure1]{1b}, our final corpus includes 48.7\% from PubMed, 31.6\% from clinicaltrials.gov, and 19.6\% from additional global registries. In addition, we collect the systematic literature review publications from PubMed and parse their references to obtain the PubMed ID of the cited clinical studies. Based on the sections of the references, we can further get the included and excluded studies from the review.

As shown in Figure~\hyperref[fig:figure1]{1a}, for PubMed-sourced records, we gather abstracts and full texts when available, then use a large language model to extract core study attributes, including drugs, conditions, phase, primary endpoints, and related information. For global registry sources, we implement structured parsers to extract comparable fields so that all records can be transformed into a unified set of study attributes. Using this unified data, we apply a sequence of post-processing steps to standardize key fields. These include drugs, conditions, adverse events, biomarkers, and outcome measurements, which are normalized through entity extraction and matching with established databases such as DrugBank~\cite{drugbank}, MedDRA~\cite{meddra}, TheMarker~\cite{10.1093/nar/gkad862}, RxNorm~\cite{RxNorm}, MeSH term~\cite{meshterm}, and COMET~\cite{kirkham2022core}. The entire pipeline integrates heuristic rules with LLM-based extraction.

As illustrated in Figure~\hyperref[fig:figure1]{1a}, the goal of the post-processing stage is to construct a unified database of clinical trials. This includes a knowledge graph that links trials through shared target drugs, conditions, and other standardized entities, as well as connections based on shared systematic reviews, which often signal domain-relatedness. We also create a standardized relational database with multiple tables that capture detailed trial information, including studies, conditions, biomarkers, endpoints, adverse events, disposition, outcomes, and results. As shown in Table~\ref{tab:benchmark_comparison}, we compare \datasetname against the most related databases from the literature and highlight its comprehensiveness in sources and the data field coverage. Figures~\hyperref[fig:figure1]{1c} and~\hyperref[fig:figure1]{1d} show the distributions of target drugs and conditions in \datasetname. We observe that a small set of agents appears with very high frequency, such as ethanol, ethinyloestradiol, paclitaxel, acetylsalicylic acid, and metformin, while the long tail consists of many agents used in only a modest number of trials. A similar pattern is present for conditions, where general categories related to pathologic processes, signs and symptoms, and broad disease classes appear most often, followed by a long tail of more specific disease groups. This unified resource enables efficient large-scale extraction and analysis of clinical trial information and serves as a strong data foundation for developing large language models that support clinical research tasks.


\subsection*{A pipeline for data extraction and LLM development}
\begin{figure}
    \centering
    \includegraphics[width=0.95\linewidth]{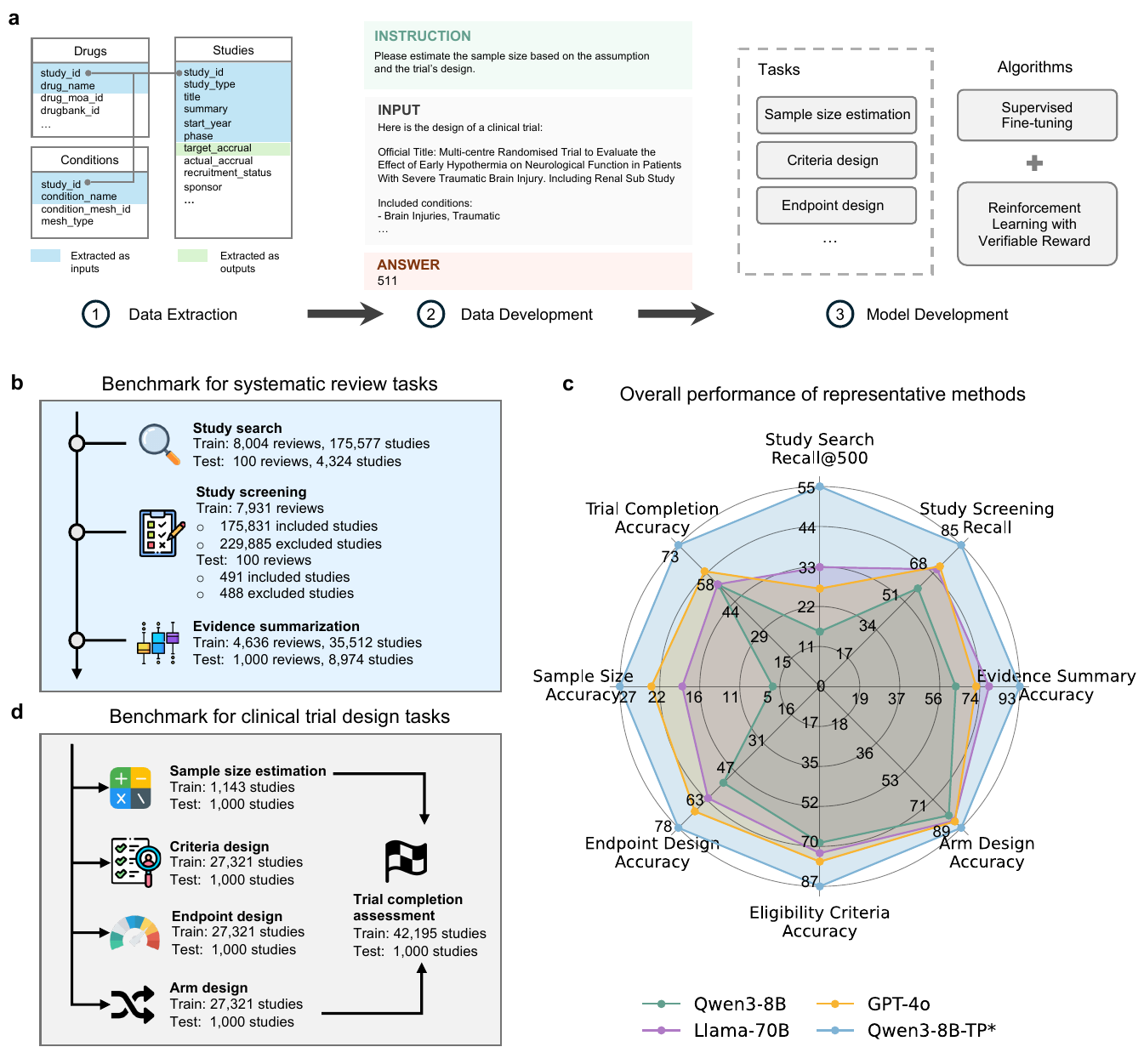}
    \caption{Benchmark construction and model performance across systematic review and clinical trial design tasks. \textbf{a}, End-to-end workflow for transforming structured and semi-structured clinical trial records into supervised training examples. Extracted study attributes are composed into standardized instruction input answer triples, enabling the development of models for tasks such as sample size estimation, criteria design, and endpoint design. \textbf{b}, Overview of the benchmark for systematic review tasks, including study search, study screening, and evidence summarization, together with the scale of training and testing sets derived from real-world review corpora. \textbf{c}, Overall performance of representative large language models on the full suite of benchmark tasks. ``Qwen3-8B-TP*'' indicates Qwen3-8B model fine-tuned using the mixture of supervised fine-tuning (SFT) and reinforcement learning with verifiable reward (RLVR) strategies on \datasetname. Results illustrate substantial variation across tasks, with domain-adapted models achieving stronger accuracy and recall in both review-oriented and design-oriented evaluations. \textbf{d}, Benchmark tasks for clinical trial design covering sample size estimation, eligibility criteria design, endpoint selection, arm design, and trial completion assessment, along with the number of training and testing instances for each task.}
    \label{fig:figure2}
\end{figure}

\datasetname enables efficient extraction of development data for large language models. Figure~\hyperref[fig:figure2]{2a} illustrates an example workflow for constructing development data for the sample size estimation task. In Step 1, we identify the study fields that serve as inputs to the task. For this example, we use study-level attributes such as study type, year, phase, condition, and related descriptors as the trial setup, and use the field target accrual as the sample size value to be predicted. We further select the studies with full text so as to extract the statistical assumptions stated in the statistical analysis plan section. In Step 2, we organize these inputs into a standardized format for model development, which includes the instruction that specifies the task, the input that provides the trial setup as context, and the answer that contains the ground truth value for supervised learning. Through this approach, we can flexibly select study fields from the relational database and the knowledge graph to create instruction data tailored to many clinical research tasks. 

As shown in Figures \hyperref[fig:figure2]{2b} and \hyperref[fig:figure2]{2d}, we use the \datasetname pipeline to extract development data for three systematic review tasks and five clinical trial design tasks. The systematic review tasks cover the core steps of a review workflow. The study search task takes a review topic as input and identifies the most relevant clinical trials from literature sources such as PubMed. The study screening task takes the review selection criteria and a candidate study set and evaluates the eligibility of each study for inclusion. The evidence summarization task takes the results of multiple clinical trials and produces a concise evidence summary that can support future clinical decision-making. For clinical trial design tasks, we include four tasks focused on study design fields: sample size estimation, criteria design, endpoint design, and arm design. We also include a trial completion assessment task that takes the trial setup as input, predicts whether the trial is likely to complete or terminate, and explains the factors that drive the prediction. Together, these tasks establish a comprehensive benchmark for large language models (LLMs) in clinical research settings and provide train-test splits for building vertical models tailored to clinical research applications. We hold out the latest studies as the test set to avoid the data leakage issues.

In Figure~\hyperref[fig:figure2]{2a}, Step 3 applies a combination of supervised fine-tuning (SFT) and reinforcement learning with verifiable reward (RLVR) to adapt generic large language models to clinical research tasks. We first train models with SFT for all tasks, then apply RLVR to tasks that require deeper reasoning and offer a verifiable reward signal, such as sample size estimation. This staged approach enables broad task coverage while strengthening reasoning ability. Building on this framework, \datasetname benchmark provides a unified testbed to evaluate both open and proprietary models and to develop domain-adapted variants. For example, we adapt Qwen3-8B to clinical research by training it on the \datasetname corpus, producing Qwen3-8B-TP. As shown in Figure~\hyperref[fig:figure2]{2c}, Qwen3-8B-TP delivers clear gains over the base Qwen3-8B across all tasks. Remarkably, this 8B vertical model surpasses Llama-70B, despite using about ten times fewer parameters, and also outperforms GPT-4o, underscoring the value of targeted post-training for clinical research workflows.

\subsection*{LLMs for clinical trial design and optimization}
\begin{figure}
    \centering
    \includegraphics[width=0.95\linewidth]{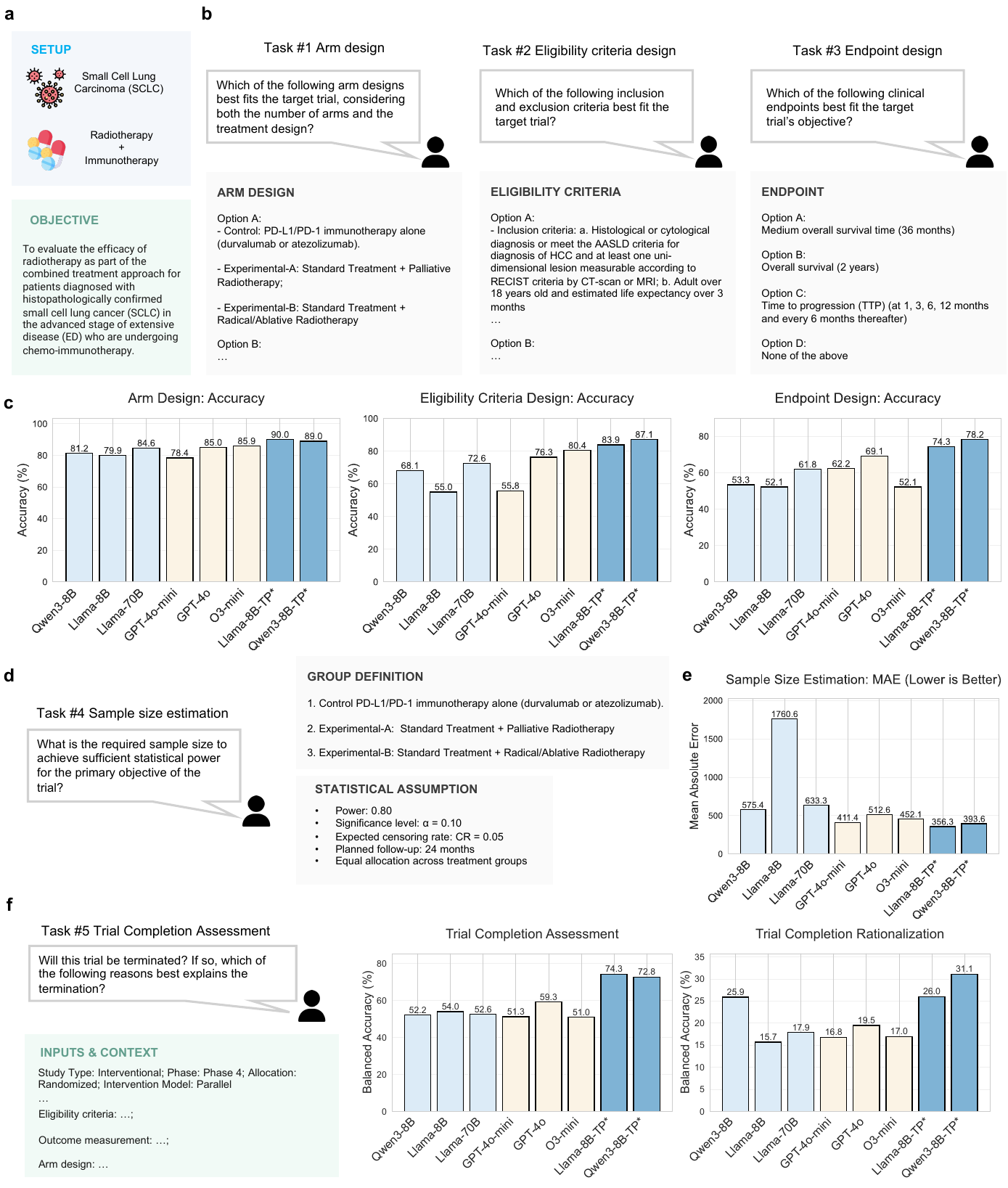}
    \caption{Benchmark tasks for clinical trial design and comparative performance of large language models. \textbf{a}, Example trial setup illustrating how key attributes, such as indication, interventions, and study objective, form the context for downstream design tasks. \textbf{b}, Illustrations of three multiple-choice design tasks: arm design, eligibility criteria design, and endpoint design, where the model must select the option that best aligns with the target trial. \textbf{c}, Accuracy of eight models on arm design, eligibility criteria design, and endpoint design, showing clear gains from domain-adapted training. \textbf{d}, Illustration of the sample size estimation task, including group definitions and statistical assumptions used to construct ground truth labels. \textbf{e}, Mean absolute error for sample size estimation across eight models, where lower error indicates better performance. \textbf{f}, Trial completion assessment task and corresponding performance, evaluated by balanced accuracy for termination prediction and rationale selection, highlighting improvements in models trained on \datasetname.}
    \label{fig:figure3}
\end{figure}

We present all results for trial design tasks in Figure~\hyperref[fig:figure3]{3}. To benchmark LLM performance on these tasks, we include open source models Qwen3-8B~\cite{yang2025qwen3}, Llama3-8B, and Llama3-70B~\cite{dubey2024llama}, together with proprietary models GPT-4o-mini, GPT-4o, and the reasoning model o3-mini. For comparison, we also develop domain-adapted LLMs trained on \datasetname using the Qwen3-8B and Llama3-8B bases, which yield Qwen3-8B-TP and Llama3-8B-TP, where ``TP'' is short for \datasetname. Figures~\hyperref[fig:figure3]{3a} and~\hyperref[fig:figure3]{3b} illustrate the construction of the three trial design tasks. Each task presents the target trial setup on the left, including the disease area, treatment modalities, and study objective, followed by a multiple-choice question that mirrors common decisions made during protocol development. For arm design, models must determine which treatment arms best match the target trial, considering both the number of arms and the treatment composition. For eligibility criteria design, models must choose the inclusion and exclusion criteria that are most appropriate for the stated objective and patient population. For endpoint design, models must select the endpoint option that aligns with the clinical goal of the trial.

Figures~\hyperref[fig:figure3]{3c} show the results of the three trial design tasks. Across arm design, eligibility criteria design, and endpoint design, proprietary frontier models outperform the open source baselines, while the domain-adapted models achieve the strongest overall accuracy. For arm design, the best open source model reaches 85.0\%, whereas GPT-4o and O3-mini reach 86.3\% and 89.0\%, respectively. The domain-adapted models further improve performance, achieving 89.0\% for Qwen3-8B-TP and 90.0\% for Llama3-8B-TP. A similar pattern is observed in eligibility criteria design, where the baseline models plateau between 55.0\% and 72.6\%, while O3-mini reaches 84.4\%. Domain adaptation again yields the highest accuracy, with Qwen3 8B TP at 87.1\%. Endpoint design shows the largest gap between open source baselines and specialized models. Baseline models remain below 62.0\%, whereas O3-mini reaches 69.1\%. The domain-adapted variants achieve 74.3\% and 78.2\%, respectively, demonstrating consistent gains from tailoring the models to the clinical trial domain. These results indicate that domain-adapted LLMs provide the most reliable performance across diverse components of trial design.

Figure~\hyperref[fig:figure3]{3d} shows the construction of the sample size estimation task, which requires multi-step statistical reasoning rather than pattern matching. The model must interpret a structured trial synopsis that includes the group definitions for control and experimental arms, along with the statistical assumptions needed for power analysis. Given this setup, the model must compute the required sample size that satisfies the primary power requirement.  We assess model performance by computing the mean absolute error (MAE) of the model-estimated sample size against the original trial's target accrual, which was made by human statisticians. As shown in Figure~\hyperref[fig:figure3]{3e}, open source models exhibit large estimation errors, with mean absolute error ranging from 575.4 to 1769.6. Proprietary frontier models reduce the error substantially, reaching 411.4 for GPT-4o-mini and 452.1 for GPT-4o. Domain-adapted models achieve the best accuracy overall, with a mean absolute error of 356.3 for Llama3-8B-TP and 393.6 for Qwen3-8B-TP. These results highlight both the difficulty of this reasoning-heavy task and the effectiveness of targeted domain adaptation for improving statistical modeling performance.

Figure~\hyperref[fig:figure3]{3f} shows the trial completion assessment task, where the model predicts whether a clinical trial will terminate early and, if so, selects the most plausible reason among options such as enrollment issues, safety concerns, lack of efficacy, or operational problems. The input includes structured trial attributes such as study type, phase, allocation, intervention model, eligibility criteria, outcome measurements, and arm design, requiring the model to integrate heterogeneous signals before making a binary prediction and generating a rationalization. As shown in the left panel, open source models achieve balanced accuracy between 52.2\% and 56.4\%, while GPT-4o reaches 59.3\%. Domain-adapted models again deliver the highest performance, with Llama3-8B-TP reaching 74.3\% and Qwen3-8B-TP reaching 72.8\%. The right panel reports the rationalization accuracy, which is substantially more challenging. Baseline models remain below 26.0\%, whereas the domain-adapted variants reach 26.0\% for Llama3-8B-TP and 31.1\% for Qwen3-8B-TP. These results indicate that domain adaptation improves both the termination prediction and the ability to identify the underlying reason, even though rationalization remains a challenging task.

\subsection*{LLMs for systematic literature review of clinical trials}
\begin{figure}
    \centering
    \includegraphics[width=0.95\linewidth]{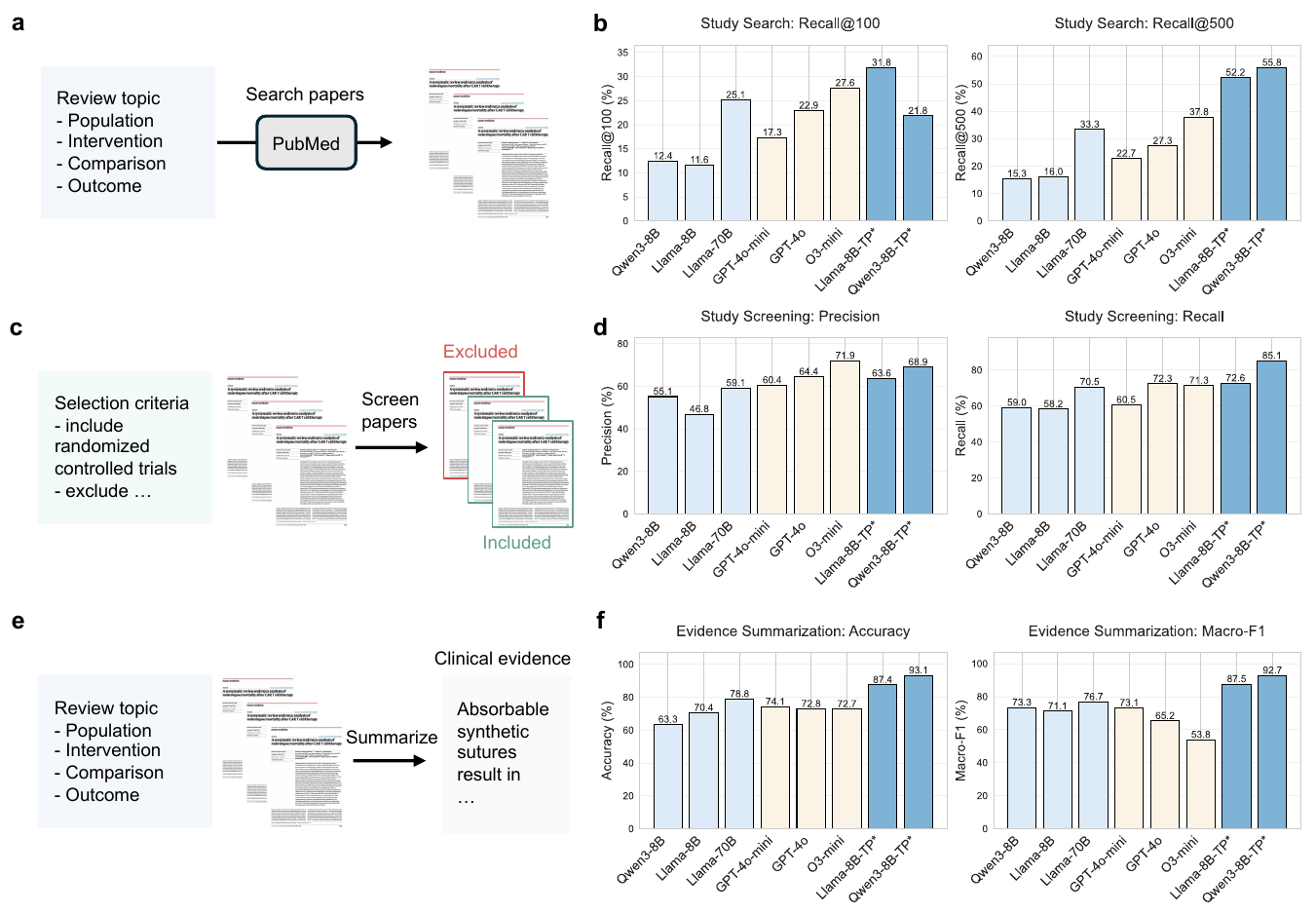}
    \caption{Systematic review task setup and model performance across study search, study screening, and evidence summarization. \textbf{a}, Study search task formulation: given a review topic specified by the PICO elements, the model must generate an effective PubMed query to retrieve relevant clinical studies. \textbf{b}, Performance of eight models on study search measured by Recall at one hundred and Recall at five hundred, highlighting substantial gains from domain-adapted training. \textbf{c}, Study screening task formulation: given review selection criteria and candidate studies, the model classifies each study as included or excluded. \textbf{d}, Precision and recall for study screening across models, showing improvements in both specificity and sensitivity with domain-adapted models. \textbf{e}, Evidence summarization task formulation: given a review topic and selected studies, the model must identify the correct evidence statement among multiple candidates or select none if no option is supported. \textbf{f}, Accuracy and Macro F1 for evidence summarization, demonstrating strong benefits of task-aligned training for fine-grained evidence reasoning.}
    \label{fig:figure4}
\end{figure}

Systematic literature review (SLR) is a foundational activity in clinical research, yet it remains one of the most labor-intensive and time-consuming stages of the trial lifecycle. In early trial planning, SLR supports evidence-grounded decisions by mapping the landscape of prior studies, identifying safety signals and adverse events in relevant patient groups, and establishing realistic benchmarks for study design. In later stages, SLR enables the integration of clinical evidence across historical trials to inform guideline development and clinical practice. To capture these essential workflows, we design a set of SLR tasks centered on three key steps: study search, study screening, and evidence summarization. These tasks, illustrated in Figure~\hyperref[fig:figure4]{4}, provide a structured benchmark for evaluating how well LLMs can assist researchers throughout the clinical evidence synthesis pipeline.

Figure~\hyperref[fig:figure4]{4a} illustrates the setup of the study search task. Given a review topic specified through the PICO framework, including the target population, intervention, comparison group, and outcome measures, the model must generate an effective search query for a literature database such as PubMed in order to retrieve as many relevant clinical studies as possible. We evaluate retrieval performance using Recall@100 and Recall@500, which measure the coverage of ground truth studies (i.e., the ones that were included in the review paper we reference in the test case) within the top retrieved results. As shown in Figure~\hyperref[fig:figure4]{4b}, proprietary models and domain-adapted models achieve markedly stronger coverage than general open source baselines. GPT-4o and GPT-4o mini outperform Qwen3 and Llama models at both retrieval depths, and the specialized models trained on our dataset Qwen3-8B-TP and Llama3-8B-TP provide further gains. Under Recall@100, Qwen3-8B-TP reaches approximately 32\% and Llama3-8B-TP reaches approximately 31\%, compared with approximately 17\% for Qwen3-8B and approximately 19\% for Llama3-8B. Under Recall@500, Qwen3-8B-TP attains approximately 57\%, and Llama3-8B-TP attains approximately 54\%, while the strongest proprietary baseline GPT-4o reaches approximately 47\%. These results show that targeted post-training on our clinical trial corpus substantially enhances an LLM's ability to generate precise and comprehensive search queries for SLR workflows.

Figure~\hyperref[fig:figure4]{4c} presents the setup of the study screening task. Given a set of papers retrieved for a review topic and the associated inclusion and exclusion criteria specified by the review, the model must decide for each paper whether it should be included or excluded. This task reflects the core filtering stage of an SLR workflow, where researchers determine which studies meet the methodological and clinical requirements of the review. We evaluate model performance using precision and recall, which together capture how accurately a model identifies eligible studies without introducing excessive false positives or false negatives. As shown in Figure~\hyperref[fig:figure4]{4d}, proprietary models again surpass general-purpose open source LLMs, and the domain-adapted models yield the strongest overall performance. O3-mini and GPT-4o show clear gains over Qwen3 and Llama baselines, but the TP variants provide the largest improvements. Llama3-8B-TP achieves a precision of approximately 72\%, and Qwen3-8B-TP reaches approximately 69\%, compared with approximately 55\% to 60\% for the base open source models. In terms of recall, Qwen3-8B-TP attains approximately 86\%, and Llama3-8B-TP reaches approximately 72\%, again exceeding both open source baselines and proprietary models. These results indicate that targeted post-training substantially improves an LLM's ability to apply complex inclusion and exclusion rules and reliably identify relevant studies for SLR workflows.

Figure~\hyperref[fig:figure4]{4e} shows the setup of the evidence summarization task. Given a review topic specified through the PICO framework and the set of studies included after screening, the model is presented with multiple candidate evidence statements and must identify which one accurately reflects the clinical finding, or determine that none of the candidates is correct. This formulation mirrors the synthesis stage in SLR workflows, where reviewers must discern which textual statements faithfully represent the aggregated evidence. We evaluate model performance using accuracy and macro F1, which jointly capture the ability to select the correct evidence statement while avoiding false selections when no option is appropriate. As shown in Figure~\hyperref[fig:figure4]{4f}, proprietary models outperform general open source baselines, and domain-adapted models again achieve the strongest results. GPT-4o and O3-mini provide clear improvements over Qwen3 and Llama models, but the TP variants show the most substantial gains. Qwen3-8B-TP reaches an accuracy of approximately 82\% and a macro F1 of approximately 81\%, the highest among all evaluated models. Llama3-8B-TP also performs well with accuracy near 76\% and macro F1 near 75\%, compared with accuracy values between 67\% and 74\% for the general-purpose baselines. These results demonstrate that targeted post-training on clinical trial data significantly enhances an LLM's ability to distinguish accurate evidence from distractors, a critical capability for reliable evidence synthesis in SLR workflows.
\section*{Discussion}
\datasetname represents a step toward enabling cutting-edge large language models (LLMs) to support clinical research, an essential component in accelerating therapy development and improving population health. Our work addresses two central challenges in building effective domain models for this setting. First, developing domain-specialized LLMs requires high-quality and large-volume vertical data, yet such resources have historically been fragmented or unavailable at scale. Second, there has been no standardized pipeline for transforming raw clinical research data into development-ready training corpora or for applying post-training strategies, including supervised finetuning (SFT) and reinforcement learning with verifiable reward (RLVR), that reliably strengthen model reasoning.

Here, we first assemble over 1.6 million clinical trial records and 30K systematic literature reviews from literature databases, ClinicalTrials.gov, and additional global trial registries. After multi-stage cleaning, harmonization, and ontology-based standardization, these sources are integrated into \datasetname, which provides a unified trial database and a knowledge graph linking review papers to clinical studies and connecting studies through shared drugs, conditions, and other biomedical entities. Using this resource, we show how developers can efficiently extract structured supervision and construct development data for training and evaluating LLMs across five trial design and optimization tasks and three systematic review tasks. Our benchmark compares eight open and proprietary models alongside \datasetname adapted LLMs. The results consistently demonstrate that domain-adapted models outperform all generic baselines. Through the combination of \datasetname and a unified SFT plus RLVR pipeline, we observe substantial gains in reasoning quality. In particular, our 8B-level models surpass a 70B generic model and a proprietary reasoning model such as o3-mini, underscoring that task-aligned data and verifiable reward signals can be more consequential than scale alone for clinical research reasoning.

This study has several limitations. First, although \datasetname integrates one million trials from fifteen global registries, coverage remains uneven, particularly for regions with limited reporting, and the resulting representativity may vary across therapeutic areas and countries. The pipeline also depends heavily on registry-level text quality for fields such as eligibility criteria and outcome descriptions. Since large-scale processing relies on a mixture of heuristics and AI-based extraction, low-quality or ambiguous text can introduce noise into task inputs or training targets. Ontology normalization mitigates variability but cannot fully resolve missing or underspecified content. Second, the evaluation setup necessarily simplifies real-world clinical research workflows. We adopt proxy metrics and task formulations to enable reproducible benchmarking, but these abstractions cannot capture the full complexity of clinical decision making or the nuances of protocol design. Similarly, post-training approaches such as RLVR are applicable only to tasks with clearly verifiable labels, leaving many clinically important tasks outside the current scope. Third, domain generalizability and temporal drift remain challenges. Registry data reflect design practices at the time of registration, which may not align with evolving standards of care or contemporary regulatory expectations. Without continual updates, models trained on historical patterns may not fully capture emerging paradigms such as decentralized, adaptive, or platform trials. Although \datasetname covers a broad range of conditions and study types, model performance may vary in underrepresented therapeutic domains or rare disease contexts.

In conclusion, this work demonstrates an end-to-end pipeline for developing LLMs tailored to clinical research, spanning data construction, knowledge integration, and model adaptation. By building on a large-scale, queryable database and combining supervised finetuning with reinforcement learning using verifiable rewards, we show that domain-aligned data and post-training strategies can meaningfully elevate model performance on core clinical research tasks. We envision that such data foundations, together with continued advances in algorithms, will play a central role in enabling reliable and impactful AI systems for biomedicine.

\clearpage

\clearpage

\clearpage

\clearpage

\clearpage

\begin{table}
\centering
\caption{
Comparison of \datasetname with prior clinical trial datasets and benchmarks. The table summarizes sample size, data sources, and coverage of key structured components, including design attributes, trial results, clinical outcomes, study publication relations, and ontology linkage. It also highlights whether each resource provides benchmark tasks for trial design or systematic review workflows, showing that \datasetname offers the most comprehensive coverage across all dimensions.
}
\label{tab:benchmark_comparison}
\resizebox{0.99\linewidth}{!}{
\begin{tabular}{lccccccccc}
\toprule
\textbf{Datasets} & \textbf{Sample size} & \textbf{Data source} & \textbf{Design} & \textbf{Result} & \textbf{Outcome} & \textbf{Relation} & \textbf{Ontology} & \textbf{Design Benchmark} & \textbf{Review Benchmark} \\
\midrule
\datasetname (Ours) & 1.6M & Multiple (15 sources) & \cmark & \cmark & \cmark & \cmark & \cmark & \cmark & \cmark \\
TOP~\cite{fu2022hint} & 10K & ClinicalTrials.gov & \xmark & \xmark & \cmark & \xmark & \xmark & \cmark & \xmark \\
CTKG~\cite{chen2022knowledge} & 8K & ClinicalTrials.gov & \cmark & \xmark & \xmark & \xmark & \cmark & \xmark & \xmark \\
ClinicalTrials.govResults~\cite{shi2024constructing} & 56K & ClinicalTrials.gov & \xmark & \cmark & \cmark & \xmark & \xmark & \xmark & \xmark \\
CTO~\cite{gao2024automatically} & 125K & ClinicalTrials.gov & \xmark & \cmark & \cmark & \xmark & \xmark & \xmark & \xmark \\
TrialBench~\cite{chen2024trialbench} & 400K & ClinicalTrials.gov & \cmark & \xmark & \cmark & \xmark & \cmark & \cmark & \xmark \\
CARE~\cite{naik2024care} & 700 pubs & PubMed & \xmark & \cmark & \cmark & \xmark & \xmark & \xmark & \xmark \\
PubMedKG~\cite{xu2024pubmed} & 480K & PubMed,ClinicalTrials.gov & \xmark & \xmark & \xmark & \cmark & \xmark & \xmark & \xmark \\
TrialReviewBench~\cite{wang2024accelerating} & 100 reviews & PubMed,ClinicalTrials.gov & \xmark & \xmark & \xmark & \cmark & \xmark & \xmark & \cmark \\
LEADSInstruct~\cite{wang2025foundation} & 21K reviews & PubMed,ClinicalTrials.gov & \xmark & \xmark & \xmark & \cmark & \xmark & \xmark & \cmark \\
\bottomrule
\end{tabular}
}
\end{table}

\clearpage

\section*{Method}

\subsection*{Data collection}
To construct our benchmark dataset, we crawled clinical trial and publication records from a diverse set of publicly accessible registries and databases. The data URLs and the date of the most recent crawl are listed in Supplementary Table~\ref{tab:data_sources}. After collection, the raw data were subjected to systematic cleaning procedures to ensure consistency and reliability across sources: (1) Intra-source de-duplication: Duplicate entries within the same registry were detected and removed using unique identifiers (such as trial registration numbers), and corroborated using additional metadata like titles. (2) Inter-source de-duplication: Clinical trials registered in multiple registries were identified via fuzzy matching on titles. Duplicate records across sources were merged into a single one. (3) Field normalization: Metadata fields such as trial status, registration date, and condition labels were standardized across sources to ensure consistent querying. (4) Missing value handling: Entries with missing or incomplete critical fields were flagged. These operations minimized noise, reduced redundancy, and created a unified and clean dataset suitable for benchmarking tasks.

\subsection*{Database building}
\datasetname\ is organized into a set of structured tables that together provide a unified view of global clinical research data. The Studies table serves as the central entry for each trial and aggregates identifiers from registries and publications, trial design attributes such as phase and sponsor, and primary text fields including titles and summaries. Several complementary tables enrich each study with standardized clinical entities. The Conditions table stores disease terms mapped to Medical Subject Headings, including both primary concepts and hierarchical ancestors. The Drugs table provides a normalized representation of pharmacologic interventions, including cleaned drug names, RxNorm and DrugBank identifiers, and approval status from major regulatory agencies. The Disposition table captures the arm structure of each trial by pairing treatment groups with intervention descriptions, while the Adverse Events table records arm-specific safety findings from ClinicalTrials.gov, including frequencies, seriousness categories, and MedDRA codes.

To support evidence extraction and comparative analysis, we constructed additional tables centered on trial findings and clinical endpoints. The Trial Results table provides PICO structured summaries of outcome data at the arm level, combining extractions from PubMed abstracts with structured results from ClinicalTrials.gov. The Trial Outcomes table assigns unified outcome types to completed and terminated studies, including positive and negative classifications and fine-grained termination reasons. The Endpoints table standardizes primary outcomes by mapping them to COMET domains and subdomains, creating a consistent representation of clinical endpoints across heterogeneous sources. The Biomarkers table links text spans from trial descriptions to curated therapeutic biomarkers from TheMarker and records biomarker name, functional class, biomarker type, and associated genes when applicable.

Finally, the Relations table encodes connections among systematic reviews, cited studies, and registry records. Each entry is a structured triple that captures include, exclude, awaiting assessment, or cross registry link relationships. Together, these tables form a comprehensive relational database that supports large-scale analysis of trial design, evidence synthesis, and clinical development patterns. Detailed procedures for extraction, normalization, and validation are reported in Supplementary Section~\ref{appx:database_building}.

\subsection*{Trial design tasks: arm design}

\paragraph{Definition \& Motivation.}
This task assesses whether the model can identify the correct treatment arms and intervention details for a given clinical trial. Based on the trial’s title and brief summary, the model must choose among four candidate descriptions the one that accurately represents the trial’s arm labels, arm types, and assigned interventions. An example instance is shown in Supplementary Table~\ref{tab:trial_arm_selection}.

\paragraph{Input \& Output.}
The model is provided the official trial title and summary, followed by four arm descriptions that list the group labels, intervention types, and associated treatments. It must return the label corresponding to the correct description.

\paragraph{Dataset Construction.}
We start with sets of up to four trial identifiers drawn from trials cited together in the same Cochrane review. For each identifier, a script extracts the arm groups and individual intervention details from our trial metadata. If fewer than four candidates are available, additional trials are added at random (with a fixed seed for reproducibility). To create challenging distractors, incorrect descriptions are drawn from the same arm and intervention information of other trials cited by that review. All items are then sorted by the numeric portion of the trial identifier and partitioned so that the latest 1,000 trials form the test set, the next 500 form validation, and the rest form training.

\paragraph{Evaluation Metrics.}
We report accuracy and macro-F1 to account for class imbalance across answer choices.

\subsection*{Trial design tasks: eligibility criteria design}\label{appx:trial_design_criteria}

\paragraph{Definition \& Motivation.}
This task measures the model’s ability to match a trial’s inclusion and exclusion criteria. Given a trial’s title and summary, the model must select, from four text options, the one that accurately lists the trial’s eligibility rules. An example instance is shown in Supplementary Table~\ref{tab:trial_eligibility_selection}.

\paragraph{Input \& Output.}
The model receives the trial title and summary along with four candidate eligibility criterion sets. It must output the correct option label.

\paragraph{Dataset Construction.}
For eligibility criteria, dataset construction follows the same approach as in Arm Design: we form candidate sets from trials cited by the same Cochrane review, extract each trial’s eligibility rules from our metadata, pad with random trials when needed, and draw distractors from other cited trials to ensure relevance. Examples are then sorted by the numeric part of the trial identifier and split into test (latest 1,000), validation (next 500), and training (remainder).

\paragraph{Evaluation Metrics.}
We report Accuracy and macro‑F1 to account for class imbalance across answer choices.

\subsection*{Trial design tasks: endpoint design}

\paragraph{Definition \& Motivation.}
This task evaluates whether the model can identify the primary outcome measures of a clinical trial. Based on the trial title and summary, the model must pick the option that lists the correct primary endpoints. An example instance is shown in Supplementary Table~\ref{tab:trial_outcome_selection}.

\paragraph{Input \& Output.}
The input consists of the trial title and summary plus four outcome descriptions that concatenate the primary measures and any associated timeframes. The model must choose the correct label.

\paragraph{Dataset Construction.}
For primary endpoints, dataset construction also follows the Arm Design template: trials cited together provide candidate sets, we extract each trial’s primary outcomes from our metadata, pad with random examples if needed, and sample distractor options from other cited trials. The resulting items are ordered by the numeric portion of the trial identifier and divided into test (latest 1,000), validation (next 500), and training splits.

\paragraph{Evaluation Metrics.}
We report Accuracy and macro‑F1 to account for class imbalance across answer choices.

\subsection*{Trial design tasks: sample size estimation}

\paragraph{Definition \& Motivation.}  
This task evaluates the model’s quantitative reasoning by asking it to estimate the number of participants required for a clinical trial based on its design and stated statistical assumptions. Success requires understanding trial objectives, translating textual assumptions (effect size, significance level, power, dropout) into a numerical enrollment target, and applying standard sample‐size formulas. An example instance is shown in Supplementary Table~\ref{tab:sample_size_question}.

\paragraph{Input \& Output.}  
As input, the model receives: a narrative description of the trial design—official title, conditions, arm groups, interventions and primary outcomes, as well as a concise summary of the statistical assumptions extracted from the protocol text (treatment effect, alpha, power, expected dropout). The model must output a single integer corresponding to the estimated sample size.

\paragraph{Dataset Construction \& Splits.}  
We linked 5,737 PubMed protocols to their NCTIDs and identified 4,315 full-text sections titled ``sample size'' together with 5,678 CT.gov records containing enrollment counts. Matching the reported CT.gov enrollment with the corresponding paper section yielded 2 143 examples where the protocol paragraph explicitly mentions the official enrollment number. For each example, we used a deterministic prompt to summarize the paragraph into (1) effect‐size assumptions, (2) statistical parameters (significance level, power, dropout rate), and (3) the true enrollment. We then generated a question by combining trial metadata with these summarized assumptions. Finally, all examples were sorted by the numeric part of the NCTID; the most recent 1,000 form the test set, the next 500 the validation set, and the remainder the training set.

\paragraph{Evaluation Metrics.}  
We report (1) the mean absolute error between predicted and actual enrollments, and (2) the proportion of trials for which the model’s estimate lies within ±20 \% of the true sample size.

\subsection*{Trial design tasks: trial completion assessment}
\paragraph{Definition \& Motivation.}
This task simulates a real-world feasibility analysis during clinical trial planning, where the model must assess whether a trial is likely to reach completion or be terminated early based on its protocol design. In addition, if a trial is predicted to terminate, the model is asked to identify the most likely reason for termination from a predefined set of categories (e.g., enrollment issues, safety concerns, lack of efficacy, operational problems). This reflects practical needs in the early design stage when sponsors assess trial viability. An example instance is shown in Supplementary Table~\ref{tab:trial_completion_example}.

\paragraph{Input \& Output.}
Each example consists of structured metadata and protocol text for a registered interventional study, including trial phase, number of arms, masking, intervention model, primary purpose, arm descriptions, and detailed eligibility criteria. The output is a binary label indicating whether the trial was completed or terminated. For trials labeled as \texttt{terminated}, the model must also predict the most likely termination reason from one of several standardized categories.

\paragraph{Dataset Construction.}
We construct the dataset using completed and terminated interventional trials from ClinicalTrials.gov with known outcomes. Structured fields are extracted from the \texttt{studies}, \texttt{disposition}, \texttt{eligibility\_criteria}, and \texttt{arms} tables in our trial database. Only trials with definitive outcome labels (i.e., \texttt{Completed} or \texttt{Terminated}) and non-missing eligibility criteria are retained. Termination reasons are normalized into broad categories using a curated taxonomy (e.g., enrollment issues, lack of efficacy, safety, sponsor decision). To avoid label leakage, we exclude any outcomes or termination reasons from the input text fields.

\paragraph{Evaluation Metrics.}
We evaluate trial completion prediction using accuracy. For terminated trials, termination reason classification is evaluated using macro-F1 to balance across categories with different frequencies.

\subsection*{Systematic literature review tasks: study search}
\paragraph{Definition \& Motivation.}
This task simulates the study search step in conducting systematic reviews, where the model must comprehend a given research question and synthesize a corresponding search query to retrieve relevant studies for inclusion. The key motivation behind this task is to streamline a traditionally labor-intensive process that typically requires the expertise of trained librarians, particularly due to the complexity and variability of search query formats across literature databases. Automating this step can significantly accelerate systematic review workflows. An example instance is presented in Supplementary Table~\ref{tab:study_search_example}.

\paragraph{Input \& Output.}
The input for this task includes the setup of a systematic review, consisting of the research background, objective, and study selection criteria. The model is expected to generate a structured search query, which is then submitted to the PubMed Search API to retrieve the top 500 results and their associated PubMed IDs. A successful output should go beyond simple keyword extraction, incorporating semantic understanding and augmentation of the input to produce a comprehensive and effective query.

\paragraph{Dataset Construction.}
We constructed the dataset using a relation table that links systematic reviews to primary studies. Specifically, we focused on instances with the relation type \texttt{include}, indicating that the study was selected for inclusion in the review. For each review, we parsed its full text to extract relevant sections and used the linked PubMed abstracts as ground-truth targets. This process resulted in a dataset of 8,104 examples derived from systematic review articles, covering more than 180,000 included PubMed studies. We reserved the most recent 100 reviews as a held-out test set.

\paragraph{Evaluation Metrics.}
We evaluate this task as a retrieval problem by measuring how many ground-truth studies appear in the top search results produced by the model-generated query. Specifically, we report {Recall@100} and {Recall@500}, reflecting the number of included studies retrieved among the top 100 and 500 PubMed results, respectively.

\subsection*{Systematic literature review tasks: study screening}

\paragraph{Definition \& Motivation.}
This task simulates the study selection step in a systematic review pipeline, where the model must assess the eligibility of candidate studies for inclusion based on the review’s predefined criteria. Given a systematic review’s background, objective, and eligibility criteria, along with a set of candidate studies retrieved from PubMed (each with a title and structured abstract), the model determines whether each study should be \texttt{included} or \texttt{excluded}. This step is often labor-intensive and requires detailed reasoning across clinical nuances and protocol designs. An example instance is shown in Supplementary Table~\ref{tab:study_screening_example}.

\paragraph{Input \& Output.}
The input to this task consists of the systematic review protocol (comprising the review’s background, objective, and selection criteria) and a list of candidate studies, each represented by its title and structured abstract. The output is a binary label (\texttt{included} or \texttt{excluded}) for each candidate study, indicating whether it meets the review’s eligibility criteria. Accurate performance requires the model to match eligibility conditions against sometimes subtle or incomplete cues in the candidate study abstracts.

\paragraph{Dataset Construction.}
We construct the dataset using reviews that have explicit inclusion and exclusion annotations for candidate studies, available from the \texttt{relations} table in our clinical trial knowledge base. For each review, we extract its background, objective, and selection criteria from the abstract, and retrieve associated candidate studies along with their labels (\texttt{included} or \texttt{excluded}) based on the relation types \texttt{include} and \texttt{exclude}. To ensure a balanced and informative evaluation, we downsample the candidate pool for each review to include a random mix of included and excluded studies, preserving diversity while controlling for class imbalance. Each instance contains up to 10 candidate studies, each with a title and structured abstract, paired with a single review protocol.

\paragraph{Evaluation Metrics.}
We evaluate model performance using three metrics: accuracy, precision, and recall. Accuracy reflects the overall proportion of correctly classified studies. Precision measures the proportion of correctly identified \texttt{included} studies among all studies predicted as included, while recall measures the proportion of truly \texttt{included} studies that were correctly identified by the model. These metrics jointly assess both correctness and completeness of model screening decisions.

\subsection*{Systematic literature review tasks: evidence summarization}

\paragraph{Definition \& Motivation.}
This task simulates the final synthesis step of a systematic review, in which the model must draw a conclusion from the included studies’ findings.  Given a self-contained multiple-choice question based on a Cochrane review’s objectives, the model selects the answer option that aligns with the review’s main conclusion. An example instance is shown in Supplementary Table~\ref{tab:evidence_summarization}.

\paragraph{Input \& Output.}
The input to the evidence summarization task consists of a self‐contained multiple‐choice question generated from a Cochrane review abstract (including its answer options A, B, C, \dots) together with the titles and full structured abstracts of all studies included in that review (ranging from 1 to 20 PMIDs). The model’s output is a single label (A, B, C, \dots) corresponding to the correct conclusion. Successfully performing this task requires the model to comprehend scientific text, aggregate evidence across multiple documents, and map study findings to the correct conclusion.

\paragraph{Dataset Construction.}
First, Cochrane review abstracts are processed by GPT-4o using a prompt that instructs the model to generate a self‐contained multiple‐choice question based on the review objectives, to produce answer options covering plausible outcomes, and to output only valid JSON with the keys \texttt{question}, \texttt{options}, and \texttt{answer}.  The prompt also forbids referencing the original review by phrases such as ``based on'' to ensure questions are fully self‐contained, and the model is run at temperature 0 to guarantee deterministic outputs.  Next, for each generated question, we retrieve the included‐study PMIDs from the review metadata and load the corresponding titles and structured abstracts from our PubMed cache, discarding any entries with zero or more than twenty included studies.  Finally, the remaining $N$ examples are sorted by numeric PMID (serving as a proxy for publication date); the most recent 1,000 examples form the \textit{test} set, the preceding 500 form the \textit{validation} set, and the remainder form the \textit{training} set.

\paragraph{Evaluation Metrics.}
We report accuracy and macro-F1 to account for class imbalance across answer choices.

\subsection*{Model development}

We developed two domain-adapted large language models (LLMs) based on LLaMA-3-8B-Instruct \citep{dubey2024llama} and Qwen3-8B \citep{yang2025qwen3}, respectively. Both models were trained following a two-stage pipeline that combines supervised fine-tuning (SFT) for domain knowledge injection and reinforcement learning with verifiable rewards (RLVR) for improving task-specific performance.

\subsubsection*{Supervised Fine-tuning (SFT)}
Supervised fine tuning was performed in two sequential phases. In the first phase, the model was trained only on the tasks with high volume datasets, specifically trial design and trial completion assessment. In the second phase, the model was trained on the remaining tasks, including sample size estimation, study search, study screening and evidence summarization. Separating these phases ensured that high volume tasks did not dominate the gradient signals, otherwise the substantially smaller tasks would struggle to be adequately learned by the model.

The model was optimized using the next token prediction objective applied only to answer tokens. Let $x$ denote the input sequence, which includes both the instruction and the question, and let $y = (y_1, \dots, y_K)$ denote the target answer sequence. The supervised fine tuning objective is
\begin{equation}
\mathcal{L}_{\mathrm{SFT}}
=
- \sum_{t=1}^{K} \log p_{\theta}\bigl( y_t \mid y_{<t}, x \bigr),
\end{equation}
so that the input $x$ only serves as conditioning context, while the loss is computed exclusively over the answer tokens $y$.

We implement SFT recipe by using the LLaMA Factory framework \citep{zheng2024llamafactory}. Training used a context length of 8192 tokens and a peak learning rate of $1\times 10^{-6}$ under a cosine decay schedule. The AdamW optimizer was employed with a batch size of 16 and numerical precision was set to bfloat16. To increase computational efficiency, DeepSpeed ZeRO-3 \citep{rajbhandari2020zero} and FlashAttention-2 \citep{dao2024flashattention} were incorporated. Phase 1 was trained for three epochs, and Phase 2 continued for an additional ten epochs.

\subsubsection*{Reinforcement Learning with Verifiable Rewards (RLVR)}

After SFT, we further optimized the models using RLVR for tasks that admit automatically verifiable outcomes, specifically sample size estimation and study search. Training followed a two-step recipe. The first step trained the model on sample size estimation, and the second step trained the model on study search. 

RLVR was implemented using the VeRL framework \citep{sheng2025hybridflow} with GRPO \citep{shao2024deepseekmath} as the underlying optimization algorithm.  
For each input $x$, we sample $G$ output sequences 
$y_i = (y_{i,1},\dots,y_{i,|y_i|})$ from the old policy 
$\pi_{\theta_{\mathrm{old}}}(\cdot \mid x)$, and compute a scalar reward 
$r_i$ for each sequence.  
The GRPO objective maximizes a clipped policy improvement term averaged across all rollouts and all tokens:

\begin{equation}
\scalebox{0.95}{$
\mathcal{J}_{\mathrm{GRPO}}(\theta)
=
\mathbb{E}_{x,\; \{y_i\}_{i=1}^{G} \sim \pi_{\theta_{\mathrm{old}}}(\cdot\mid x)}
\left[
\frac{1}{G|y_i|}
\sum_{i=1}^{G}
\sum_{t=1}^{|y_i|}
\min\!\left(
\frac{
\pi_{\theta}(y_{i,t}\mid x, y_{i,<t})
}{
\pi_{\theta_{\mathrm{old}}}(y_{i,t}\mid x, y_{i,<t})
}
A_i,\;
\operatorname{clip}\!\left(
\frac{
\pi_{\theta}(y_{i,t}\mid x, y_{i,<t})
}{
\pi_{\theta_{\mathrm{old}}}(y_{i,t}\mid x, y_{i,<t})
},
1-\varepsilon,\;
1+\varepsilon
\right)
A_i
\right)
\right]
$}
\end{equation}
where the scalar advantage $A_i$ for rollout $i$ is computed by normalizing rewards within the group of $G$ rollouts,

\begin{equation}
A_i
=
\frac{
r_i - \mathrm{mean}(\{r_1,\dots,r_G\})
}{
\mathrm{std}(\{r_1,\dots,r_G\})
}.
\end{equation}

Training used a prompt length limit of 2,048 tokens, a response length limit of 512 tokens, batch sizes of 16 for training and 32 for validation, and a learning rate of $1\times10^{-6}$. During rollout generation, the sampling temperature was set to 0.6, nucleus sampling used a top-p of 0.95, and eight rollouts were generated per prompt. Each RLVR stage was trained for five epochs.

Across both tasks, rewards combined a format reward and a performance reward. The format reward enforced adherence to the required output structure. Specifically, the model was required to present its final answer within an <answer> ... </answer> tag pair. Outputs that failed to satisfy this requirement received a penalty of $-2$. When the answer was correctly enclosed within the prescribed tags, the format reward was set to $0.1$. Conditional on correct formatting, the performance reward evaluated task accuracy. For the study search task, performance was defined by retrieval quality, measured using recall$@100$. For the sample size estimation task, performance was determined by whether the predicted sample size fell within a tolerance window between 0.8 and 
1.2 times of the ground truth. Further, if the performance reward was zero or negative, the formatting bonus was not applied. This reward structure ensures that correctly formatted but substantively incorrect outputs do not accumulate positive reward, thereby aligning the learning signal with the practical expectations.

The final models, LLaMA-3-8B-TP and Qwen-3-8B-TP, were produced by applying the supervised fine-tuning and reinforcement learning pipeline described above. These models exhibit substantially improved performance across trial design tasks and systematic review tasks compared with their base versions, indicating the importance of domain aligned post training strategies.

\section*{Data Availability}
The database is available at \url{https://huggingface.co/datasets/zifeng-ai/TrialPanorama-database}. The built training and evaluation datasets are available at \url{https://huggingface.co/datasets/zifeng-ai/TrialPanorama-benchmark}.

\section*{Code Availability}
The code for experiments used in this study is available at \url{https://github.com/RyanWangZf/TrialPanorama}.

\bibliographystyle{naturemag}
\bibliography{main}

@article{shi2024constructing,
  title={Constructing a finer-grained representation of clinical trial results from ClinicalTrials. gov},
  author={Shi, Xuanyu and Du, Jian},
  journal={Scientific Data},
  volume={11},
  number={1},
  pages={41},
  year={2024},
  publisher={Nature Publishing Group UK London}
}

@article{chen2024trialbench,
  title={Trialbench: Multi-modal artificial intelligence-ready clinical trial datasets},
  author={Chen, Jintai and Hu, Yaojun and Wang, Yue and Lu, Yingzhou and Cao, Xu and Lin, Miao and Xu, Hongxia and Wu, Jian and Xiao, Cao and Sun, Jimeng and others},
  journal={arXiv preprint arXiv:2407.00631},
  year={2024}
}

@article{gao2024automatically,
  title={Automatically Labeling \$200 B Life-Saving Datasets: A Large Clinical Trial Outcome Benchmark},
  author={Gao, Chufan and Pradeepkumar, Jathurshan and Das, Trisha and Thati, Shivashankar and Sun, Jimeng},
  journal={arXiv preprint arXiv:2406.10292},
  year={2024}
}

@article{avram2023drugcentral,
  title={DrugCentral 2023 extends human clinical data and integrates veterinary drugs},
  author={Avram, Sorin and Wilson, Thomas B and Curpan, Ramona and Halip, Liliana and Borota, Ana and Bora, Alina and Bologa, Cristian G and Holmes, Jayme and Knockel, Jeffrey and Yang, Jeremy J and others},
  journal={Nucleic acids research},
  volume={51},
  number={D1},
  pages={D1276--D1287},
  year={2023},
  publisher={Oxford University Press}
}

@article{fu2022hint,
  title={Hint: Hierarchical interaction network for clinical-trial-outcome predictions},
  author={Fu, Tianfan and Huang, Kexin and Xiao, Cao and Glass, Lucas M and Sun, Jimeng},
  journal={Patterns},
  volume={3},
  number={4},
  year={2022},
  publisher={Elsevier}
}

@article{chen2022knowledge,
  title={A knowledge graph of clinical trials (CTKG)},
  author={Chen, Ziqi and Peng, Bo and Ioannidis, Vassilis N and Li, Mufei and Karypis, George and Ning, Xia},
  journal={Scientific reports},
  volume={12},
  number={1},
  pages={4724},
  year={2022},
  publisher={Nature Publishing Group UK London}
}

@inproceedings{naik2024care,
  title={CARE: Extracting Experimental Findings From Clinical Literature},
  author={Naik, Aakanksha and Kuehl, Bailey and Bransom, Erin and Downey, Doug and Hope, Tom},
  booktitle={Findings of the Association for Computational Linguistics: NAACL 2024},
  pages={4580--4596},
  year={2024}
}

@article{xu2024pubmed,
  title={PubMed knowledge graph 2.0: Connecting papers, patents, and clinical trials in biomedical science},
  author={Xu, Jian and Yu, Chao and Xu, Jiawei and Ding, Ying and Torvik, Vetle I and Kang, Jaewoo and Sung, Mujeen and Song, Min},
  journal={arXiv preprint arXiv:2410.07969},
  year={2024}
}

@article{martin2017much,
  title={How much do clinical trials cost?},
  author={Martin, Linda and Hutchens, Melissa and Hawkins, Conrad and Radnov, Alaina},
  journal={Nature Reviews Drug Discovery},
  volume={16},
  number={6},
  pages={381--382},
  year={2017},
  publisher={Nature Publishing Group UK London}
}

@article{xu2020building,
  title={Building a PubMed knowledge graph},
  author={Xu, Jian and Kim, Sunkyu and Song, Min and Jeong, Minbyul and Kim, Donghyeon and Kang, Jaewoo and Rousseau, Justin F and Li, Xin and Xu, Weijia and Torvik, Vetle I and others},
  journal={Scientific data},
  volume={7},
  number={1},
  pages={205},
  year={2020},
  publisher={Nature Publishing Group UK London}
}

@article{brbic2024predicting,
  title={Predicting drug outcome of population via clinical knowledge graph},
  author={Brbi{\'c}, Maria and Yasunaga, Michihiro and Agarwal, Prabhat and Leskovec, Jure},
  journal={medRxiv},
  year={2024}
}

@article{field2010meta,
  title={How to do a meta-analysis},
  author={Field, Andy P and Gillett, Raphael},
  journal={British Journal of Mathematical and Statistical Psychology},
  volume={63},
  number={3},
  pages={665--694},
  year={2010},
  publisher={Wiley Online Library}
}

@incollection{concato2017randomized,
  title={Randomized, controlled trials, observational studies, and the hierarchy of research designs},
  author={Concato, John and Shah, Nirav and Horwitz, Ralph I},
  booktitle={Research Ethics},
  pages={207--212},
  year={2017},
  publisher={Routledge}
}

@misc{clinicaltrialsgov,
title={ClinicalTrials.gov is a place to learn about clinical studies from around the world.},
howpublished={\url{https://clinicaltrials.gov/}},
note = {Accessed: 2025-05-01}
}

@misc{whoictrp,
title={International Clinical Trials Registry Platform (ICTRP)},
howpublished={\url{https://www.who.int/tools/clinical-trials-registry-platform}},
note = {Accessed: 2025-05-01}
}

@misc{euclinicaltrial,
title={EU Clinical Trials Register},
howpublished={\url{https://www.clinicaltrialsregister.eu/about.html}},
note = {Accessed: 2025-05-01}
}

@misc{chinesechictr,
title={Chinese Clinical Trial Registry},
howpublished={\url{https://www.chictr.org.cn/indexEN.html}},
note = {Accessed: 2025-05-01}
}

@misc{anzctr,
title={Australian New Zealand Clinical Trials Registry},
howpublished={\url{https://www.anzctr.org.au/}},
note = {Accessed: 2025-05-01}
}

@inproceedings{wang2023spot,
  title={Spot: Sequential predictive modeling of clinical trial outcome with meta-learning},
  author={Wang, Zifeng and Xiao, Cao and Sun, Jimeng},
  booktitle={Proceedings of the 14th ACM International Conference on Bioinformatics, Computational Biology, and Health Informatics},
  pages={1--11},
  year={2023},
}

@article{jin2024matching,
  title={Matching patients to clinical trials with large language models},
  author={Jin, Qiao and Wang, Zifeng and Floudas, Charalampos S and Chen, Fangyuan and Gong, Changlin and Bracken-Clarke, Dara and Xue, Elisabetta and Yang, Yifan and Sun, Jimeng and Lu, Zhiyong},
  journal={Nature Communications},
  volume={15},
  number={1},
  pages={9074},
  year={2024},
  publisher={Nature Publishing Group UK London}
}

@inproceedings{wang2023autotrial,
  title={AutoTrial: Prompting Language Models for Clinical Trial Design},
  author={Wang, Zifeng and Xiao, Cao and Sun, Jimeng},
  booktitle={Proceedings of the 2023 Conference on Empirical Methods in Natural Language Processing},
  pages={12461--12472},
  year={2023}
}

@article{wang2025informgen,
    author = {Wang, Zifeng and Gao, Junyi and Danek, Benjamin and Theodorou, Brandon and Shaik, Ruba and Thati, Shivashankar and Won, Seunghyun and Sun, Jimeng},
    title = {Compliance and factuality of large language models for clinical research document generation},
    journal = {Journal of the American Medical Informatics Association},
    pages = {ocaf174},
    year = {2025},
    month = {10},
    issn = {1527-974X},
    doi = {10.1093/jamia/ocaf174},
    url = {https://doi.org/10.1093/jamia/ocaf174},
}

@article{wang2025foundation,
  title={A foundation model for human-ai collaboration in medical literature mining},
  author={Wang, Zifeng and Cao, Lang and Jin, Qiao and Chan, Joey and Wan, Nicholas and Afzali, Behdad and Cho, Hyun-Jin and Choi, Chang-In and Emamverdi, Mehdi and Gill, Manjot K and others},
  journal={Nature Communications},
  volume={16},
  number={1},
  pages={8361},
  year={2025},
  publisher={Nature Publishing Group UK London}
}

@article{wang2024accelerating,
  title={Accelerating clinical evidence synthesis with large language models},
  author={Wang, Zifeng and Cao, Lang and Danek, Benjamin and Jin, Qiao and Lu, Zhiyong and Sun, Jimeng},
  journal={npj Digital Medicine},
  volume={8},
  number={1},
  pages={509},
  year={2025},
  publisher={Nature Publishing Group UK London}
}

@article{yuan2019criteria2query,
  title={{Criteria2Query}: a natural language interface to clinical databases for cohort definition},
  author={Yuan, Chi and Ryan, Patrick B and Ta, Casey and Guo, Yixuan and Li, Ziran and Hardin, Jill and Makadia, Rupa and Jin, Peng and Shang, Ning and Kang, Tian and others},
  journal={Journal of the American Medical Informatics Association},
  volume={26},
  number={4},
  pages={294--305},
  year={2019},
  publisher={Oxford University Press}
}

@article{park2024criteria2query,
  title={{Criteria2Query} 3.0: Leveraging generative large language models for clinical trial eligibility query generation},
  author={Park, Jimyung and Fang, Yilu and Ta, Casey and Zhang, Gongbo and Idnay, Betina and Chen, Fangyi and Feng, David and Shyu, Rebecca and Gordon, Emily R and Spotnitz, Matthew and others},
  journal={Journal of Biomedical Informatics},
  volume={154},
  pages={104649},
  year={2024},
  publisher={Elsevier}
}

@article{wornow2025zero,
  title={Zero-shot clinical trial patient matching with llms},
  author={Wornow, Michael and Lozano, Alejandro and Dash, Dev and Jindal, Jenelle and Mahaffey, Kenneth W and Shah, Nigam H},
  journal={NEJM AI},
  volume={2},
  number={1},
  pages={AIcs2400360},
  year={2025},
  publisher={Massachusetts Medical Society}
}

@article{dunn2020time,
  title={Is it time for computable evidence synthesis?},
  author={Dunn, Adam G and Bourgeois, Florence T},
  journal={Journal of the American Medical Informatics Association},
  volume={27},
  number={6},
  pages={972--975},
  year={2020},
  publisher={Oxford University Press}
}

@article{holzhauer2023super,
  title={Super-covariates: Using predicted control group outcome as a covariate in randomized clinical trials},
  author={Holzhauer, Bj{\"o}rn and Adewuyi, Emmanuel Taiwo},
  journal={Pharmaceutical Statistics},
  volume={22},
  number={6},
  pages={1062--1075},
  year={2023},
  publisher={Wiley Online Library}
}

@article{kim2017broadening,
  title={Broadening eligibility criteria to make clinical trials more representative: American Society of Clinical Oncology and Friends of Cancer Research Joint Research Statement},
  author={Kim, Edward S and Bruinooge, Suanna S and Roberts, Samantha and Ison, Gwynn and Lin, Nancy U and Gore, Lia and Uldrick, Thomas S and Lichtman, Stuart M and Roach, Nancy and Beaver, Julia A and others},
  journal={Journal of Clinical Oncology},
  volume={35},
  number={33},
  pages={3737--3744},
  year={2017},
  publisher={American Society of Clinical Oncology}
}

@inproceedings{jin2020predicting,
  title={Predicting Clinical Trial Results by Implicit Evidence Integration},
  author={Jin, Qiao and Tan, Chuanqi and Chen, Mosha and Liu, Xiaozhong and Huang, Songfang},
  booktitle={Proceedings of the 2020 Conference on Empirical Methods in Natural Language Processing (EMNLP)},
  pages={1461--1477},
  year={2020}
}

@article{10.1093/nar/gkad862,
    author = {Zhang, Yintao and Zhou, Ying and Zhou, Yuan and Yu, Xinyuan and Shen, Xinyi and Hong, Yanfeng and Zhang, Yuxin and Wang, Shanshan and Mou, Minjie and Zhang, Jinsong and Tao, Lin and Gao, Jianqing and Qiu, Yunqing and Chen, Yuzong and Zhu, Feng},
    title = {TheMarker: a comprehensive database of therapeutic biomarkers},
    journal = {Nucleic Acids Research},
    volume = {52},
    number = {D1},
    pages = {D1450-D1464},
    year = {2023},
    month = {10},
    issn = {0305-1048},
    doi = {10.1093/nar/gkad862},
    url = {https://doi.org/10.1093/nar/gkad862},
    eprint = {https://academic.oup.com/nar/article-pdf/52/D1/D1450/55040193/gkad862.pdf},
}

@article{kirkham2022core,
  title={Core outcome sets in medical research},
  author={Kirkham, Jamie J and Williamson, Paula},
  journal={BMJ medicine},
  volume={1},
  number={1},
  pages={e000284},
  year={2022}
}

@article{dodd2018taxonomy,
  title={A taxonomy has been developed for outcomes in medical research to help improve knowledge discovery},
  author={Dodd, Susanna and Clarke, Mike and Becker, Lorne and Mavergames, Chris and Fish, Rebecca and Williamson, Paula R},
  journal={Journal of Clinical Epidemiology},
  volume={96},
  pages={84--92},
  year={2018},
  publisher={Elsevier}
}

@article{dubey2024llama,
  title={The llama 3 herd of models},
  author={Dubey, Abhimanyu and Jauhri, Abhinav and Pandey, Abhinav and Kadian, Abhishek and Al-Dahle, Ahmad and Letman, Aiesha and Mathur, Akhil and Schelten, Alan and Yang, Amy and Fan, Angela and others},
  journal={arXiv e-prints},
  pages={arXiv--2407},
  year={2024}
}

@article{yang2025qwen3,
  title={Qwen3 technical report},
  author={Yang, An and Li, Anfeng and Yang, Baosong and Zhang, Beichen and Hui, Binyuan and Zheng, Bo and Yu, Bowen and Gao, Chang and Huang, Chengen and Lv, Chenxu and others},
  journal={arXiv preprint arXiv:2505.09388},
  year={2025}
}

@misc{meddra,
title={MedDRA: Medical Dictionary for Regulatory Activities},
howpublished={\url{https://www.meddra.org/}},
note = {Accessed: 2025-12-09}
}

@misc{drugbank,
title={The New DrugBank: Powered by AI},
howpublished={\url{https://go.drugbank.com/}},
note = {Accessed: 2025-12-09}
}

@misc{RxNorm,
title={Unified Medical Language System (UMLS)},
howpublished={\url{https://www.nlm.nih.gov/research/umls/rxnorm/index.html}},
note = {Accessed: 2025-12-09}
}

@misc{meshterm,
title={Welcome to Medical Subject Headings},
howpublished={\url{https://www.nlm.nih.gov/mesh/meshhome.html}},
note = {Accessed: 2025-12-09}
}

@inproceedings{rajbhandari2020zero,
  title={Zero: Memory optimizations toward training trillion parameter models},
  author={Rajbhandari, Samyam and Rasley, Jeff and Ruwase, Olatunji and He, Yuxiong},
  booktitle={SC20: International Conference for High Performance Computing, Networking, Storage and Analysis},
  pages={1--16},
  year={2020},
  organization={IEEE}
}

@inproceedings{
dao2024flashattention,
title={FlashAttention-2: Faster Attention with Better Parallelism and Work Partitioning},
author={Tri Dao},
booktitle={The Twelfth International Conference on Learning Representations},
year={2024},
url={https://openreview.net/forum?id=mZn2Xyh9Ec}
}

@inproceedings{zheng2024llamafactory,
  title={LlamaFactory: Unified Efficient Fine-Tuning of 100+ Language Models},
  author={Zheng, Yaowei and Zhang, Richong and Zhang, Junhao and YeYanhan, YeYanhan and Luo, Zheyan},
  booktitle={Proceedings of the 62nd Annual Meeting of the Association for Computational Linguistics (Volume 3: System Demonstrations)},
  pages={400--410},
  year={2024}
}

@inproceedings{sheng2025hybridflow,
  title={Hybridflow: A flexible and efficient rlhf framework},
  author={Sheng, Guangming and Zhang, Chi and Ye, Zilingfeng and Wu, Xibin and Zhang, Wang and Zhang, Ru and Peng, Yanghua and Lin, Haibin and Wu, Chuan},
  booktitle={Proceedings of the Twentieth European Conference on Computer Systems},
  pages={1279--1297},
  year={2025}
}

@article{shao2024deepseekmath,
  title={Deepseekmath: Pushing the limits of mathematical reasoning in open language models},
  author={Shao, Zhihong and Wang, Peiyi and Zhu, Qihao and Xu, Runxin and Song, Junxiao and Bi, Xiao and Zhang, Haowei and Zhang, Mingchuan and Li, YK and Wu, Yang and others},
  journal={arXiv preprint arXiv:2402.03300},
  year={2024}
}

\clearpage

\newpage
\DoToC
\newpage

\captionsetup[table]{name=Supplementary Table}
\setcounter{figure}{0}
\setcounter{table}{0}
\renewcommand*{\figurename}{Supplementary Figure}

\section{Database building}\label{appx:database_building}

\subsection{Studies table}
There are three main sources of clinical trial records in our dataset: ClinicalTrials.gov, global registries, and PubMed. ClinicalTrials.gov records are highly structured, and we extracted key attributes such as trial type, sponsor, start year, and phase from the AACT database~\footnote{\url{https://aact.ctti-clinicaltrials.org/}}.

In contrast, trial data from global registries are more heterogeneous and often semi-structured. Taking the processing of ANZCTR records as an example, we manually curated and standardized each trial entry to align with our unified schema. First, we used the "Registration number" column as the unique identifier for each study and assigned it as the `study\_id`. Trial results were extracted from the "Results – plain English summary" field and, when available, supplemented with citation links from "Citation/DOI/link/details." We also processed structured eligibility information: for instance, the gender field was standardized by mapping values like "Both males and females" to a normalized label "MALE/FEMALE," ensuring consistency across sources. Age eligibility was retained from the "Minimum age" and "Maximum age" fields, and healthy volunteer status was preserved from the corresponding binary field. Trial phases were parsed by checking substrings within the "Phase" column and converting them into discrete labels such as `PHASE1`, `PHASE2`, etc. Additionally, we extracted outcome information from the "Primary outcome" and "Secondary outcome" fields and abstracted trial objectives from the ``Brief summary'' column. These extracted attributes were compiled into a standardized table that serves as the ANZCTR subset of the merged dataset.

Moreover, clinical trials from publications like PubMed are essentially unstructured, typically presented as free-text abstracts without standardized fields. To extract structured information, we developed a rule-based pipeline that leverages regular expressions to identify key attributes embedded in the abstract. For each paper, we first extract the PubMed ID as the unique `study\_id` and retain the abstract and title as primary textual content. We then parse trial phases by searching for standard mentions such as “Phase I” or “Phase 3b” across both title and abstract, and map these to a normalized phase label (e.g., `PHASE1`, `PHASE2`, etc.). To approximate sample size, we detect phrases like “XX participants” or “XX patients” and extract the numerical value as the actual accrual count. Additionally, we identify trial results by aggregating sentences prefixed with sections such as “Results:” or “Conclusions:” when available in the abstract. This approach allows us to recover basic outcome information despite the absence of structured metadata. While this method does not capture the full complexity of trial protocols, it provides a scalable way to convert a large volume of unstructured publications into structured records suitable for downstream analysis.

\subsection{Conditions table}
To standardize disease conditions associated with each trial, we extracted Medical Subject Headings (MeSH) terms using the NCBO BioPortal Annotator API\footnote{\url{https://bioportal.bioontology.org/annotator}}. This process was applied only to non–ClinicalTrials.gov sources (e.g., PubMed and international registries), as ClinicalTrials.gov conditions are already standardized. We used the official title and brief summary of each trial as input to retrieve disease-related MeSH terms (semantic type T047) and their hierarchical ancestors. The resulting MeSH types were labeled as mesh-list (primary condition), mesh-ancestor (intermediate ancestors), and mesh-ancestor-main (top-level ancestor).

\subsection{Drugs Table}
\label{subsec:drugs_table}

To construct a comprehensive drugs table, we extract intervention mentions from three primary sources: (1) ClinicalTrials.gov and (2) PubMed. For ClinicalTrials.gov, we include entries where \texttt{intervention\_type} is labeled as \texttt{DRUG}. For PubMed, we extract the intervention component (``I'') from PICO triplets and apply a lexicon-based filter to retain only drug-related mentions, excluding non-drug interventions such as procedures, devices, or behavioral treatments.

All extracted raw mentions are then processed through a multi-stage drug normalization and mapping pipeline (Figure~\ref{fig:drug_standard}). The pipeline begins with name normalization: (1) removing common treatment-related suffixes such as “therapy” or “tablet”, (2) cleaning noisy strings by removing typographic artifacts, and (3) handling special cases including abbreviations and regional variants.

Next, drug names undergo RxNorm resolution via a dual approach combining local and remote resources. (1) Local mapping attempts to match names against a pre-indexed RxNorm database with curated mappings. (2) If local matching fails, we query the official RxNorm API, parse the returned XML, and extract the corresponding RXCUI. (3) If both methods fail, we request spelling suggestions from the API and retry mapping using the suggested variants.

For drug names successfully mapped to RxNorm, we proceed to DrugBank resolution. (1) We first attempt to map the resolved RXCUI to a DrugBank ID via a curated RxNorm-to-DrugBank mapping table. (2) If that fails, or if RxNorm mapping was unsuccessful, we apply a direct DrugBank matching routine using exact string matching, synonym expansion, brand name resolution, and fuzzy matching based on character edit distance.

This process results in three possible outcomes: (1) drug names mapped to both RxNorm and DrugBank, (2) drug names mapped to DrugBank only, and (3) drug names that remain unmapped after all resolution steps.

Table~\ref{tab:drug_mapping_stats} summarizes the drug mapping statistics for each data source, including total unique drug mentions, RxNorm mapping counts and rates, and DrugBank resolution results.

To further enrich the standardized drug table, we annotate each drug entry with regulatory approval status. Specifically, we compile lists of drugs approved by three major regulatory agencies: the U.S. Food and Drug Administration (FDA), the European Medicines Agency (EMA), and Japan’s Pharmaceuticals and Medical Devices Agency (PMDA). These lists are sourced from DrugCentral 2023~\cite{avram2023drugcentral}\footnote{\url{https://drugcentral.org/}}, which provides a consolidated view of approval data across jurisdictions.

Approval annotations are applied by checking each drug's normalized name, RxNorm-mapped name, and DrugBank-mapped name against the cleaned lists of approved drug names. A drug is marked as approved by a given agency if any of these name variants match an entry in the corresponding list. This process enables us to systematically track the clinical regulatory status of drugs within our dataset, aiding downstream tasks that require distinguishing between approved and experimental interventions.

\subsection{Trial results table}
To construct the Trial Result Table, a core component for the evidence summarization benchmark task, we systematically extracted key findings from two primary types of sources within our \datasetname~database: 1) published clinical trial literature, encompassing abstracts from PubMed, and 2) structured trial data directly from ClinicalTrials.gov registry entries. For all sources, the extraction process focused on identifying the core PICO (Population, Intervention, Comparator, Outcome) elements.

For published trial literature, a large language model (LLM), gpt-4o-mini, automatically extracted PICO components from PubMed abstracts. For data from ClinicalTrials.gov, a custom rule-based parser processed the structured JSON trial records, focusing on those with reported results (i.e., resultsSection present and outcome measures having a reportingStatus of 'POSTED'). The extracted trial results include:
\begin{itemize}
\item \textbf{P (Patient or problem):} Study population, conditions, and characteristics.
\item \textbf{N (Number of subjects):} Total participants in the population group.
\item \textbf{Outcomes (per arm):} For each reported outcome, the associated intervention arm (\textbf{I}), its status as a control (\textbf{C}, boolean flag), and detailed outcome results (\textbf{O}), including quantitative data and conclusions.
\end{itemize}
This process yielded a consistent, structured output for each abstract and document from ClinicalTrials.gov. Information extracted from both published literature and ClinicalTrials.gov was harmonized and consolidated into a unified Trial Result Table. Each row in this table represents a specific outcome from a particular trial arm. The prompt for LLM extraction is shown below.

\begin{lstlisting}
"""
Given the following clinical trial paper abstract, your task is to extract the key trial information using the PICO framework. The PICO framework consists of the following components:
Patient or problem (P): Who is the patient or what is the population or condition being addressed? Report detailed information including the population size, condition, and any relevant characteristics.
Intervention or control (I/C): What is the main intervention, treatment, or control being considered? Report detailed information including the intervention, including dosage and duration if applicable.
Outcome(s) (O): Extract detailed results and conclusions related to the outcomes, including numbers, p-values, and outcome-related and statistical conclusions if applicable.

Please return a list of reported outcome measures. Each reported outcome item should include the intervention (I) or Control (C) used and the detailed corresponding outcomes (O) of the intervention / control. Flag the control group if applicable. Only extract the information that is explicitly mentioned in the abstract. If any of the components are not mentioned in the abstract, please indicate that they are not available.

Title: {title}
Abstract: {abstract}
"""
\end{lstlisting}

\subsection{Trial outcomes table}

To extract trial outcomes, we developed two distinct workflows for trials registered on ClinicalTrials.gov and those from other international sources. For ClinicalTrials.gov trials, we first categorized them based on overall status into: (1) completed, (2) terminated/withdrawn/suspended, and (3) other (e.g., recruiting, not active, withheld). We focused on completed and terminated trials to derive fine-grained outcome types using multiple sources, including publications and reported results, as shown in Supplementary Figure~\ref{fig:outcome}.

For completed trials, if explicitly labeled as \emph{approved for marketing}, this outcome type was directly recorded. For the remaining trials, we identified matched publications by selecting the top two PubMed abstracts most relevant to the trial’s official title using text similarity. We also extracted primary outcome measures from ClinicalTrials.gov. To determine whether a trial had a positive or negative outcome, we used \emph{GPT-4o} to infer if the primary endpoint was met, based on the trial title, matched abstracts, and outcome measures. For trials without relevant abstracts or unclear outcomes, we used reported p-values for primary outcomes, applying two thresholds (0.01 and 0.05) to assign outcome types. The prompt for outcome type extraction as follows:
\begin{lstlisting}
"""
TASK:
You are a clinical trial expert. You are given a clinical trial with NCT ID [nct_id] and its official title: [official_title].

The primary outcomes of the trial are: [primary_outcomes]. 

Based on the related articles of the trial given below, please determine if the trial is likely to have a positive or negative outcome.
A positive outcome means that the trial has achieved any of its primary endpoints or outcomes. Negative outcome means that the trial has not achieved any of its primary endpoints or outcomes.
If the outcome is unclear or unable to be inferred from the articles below, please choose "unknown".

Articles:
1. Title: [top_1_similar_article_title]
Abstract: [top_1_similar_article_abstract]

1. Title: [top_2_similar_article_title]
Abstract: [top_2_similar_article_abstract]

OUTPUT:
Please provide a short answer by choosing one of the following options:
positive outcome
negative outcome
unknown

Do not provide any explanation or additional information.
"""
\end{lstlisting}

To derive fine-grained outcome types for terminated trials, we first embedded the stop reasons and visualized their clusters to identify representative keywords, which were grouped into predefined outcome types (Supplementary Figure~\ref{fig:outcome}). We further enriched these keywords using an LLM. For each trial, we computed similarity scores between the reported reason and the enriched keyword sets using a cross-encoder (ms-marco-MiniLM-L-12-v2), and assigned the outcome type with the highest average similarity. For trials from non-U.S. registries, we filtered those with reported results, which were in unstructured text format. We provided the trial title, primary outcome measures, and result text to \emph{GPT-4o} to infer the trial outcome.

\subsection{Relations table}

We implemented a multi-stage data processing pipeline centered on Cochrane Library reviews and their linked PubMed metadata to construct structured relationships between systematic reviews, clinical trials, and associated publications. Each review was first matched to its corresponding PubMed ID, and the full-text XML was parsed to extract structured reference sections. Using a custom reference extraction function, we identified citations grouped under section titles such as \textit{references to studies included in this review}, \textit{excluded from this review}, and \textit{awaiting assessment}. For each cited study, we extracted the PubMed ID and, when available, trial registry identifiers such as NCT numbers using regular expressions applied to the citation text. We formalized these relationships as triples in the format \texttt{(head\_id, relation\_type, tail\_id)}, where \texttt{head\_id} is the PubMed ID of the review, \texttt{tail\_id} is the PubMed ID of the referenced study, and \texttt{relation\_type} is one of \texttt{include}, \texttt{exclude}, or \texttt{cite}, denoting the review's assessment of the study.

We further enriched these relationships by linking PubMed studies to ClinicalTrials.gov trial records. This linkage was established through two mechanisms: first, by inspecting the \texttt{AccessionNumber} field in the PubMed XML to identify embedded NCT numbers; and second, by applying regular expressions to the abstract text to extract any mentioned NCT identifiers. In addition, we extended this mapping to include cross-registry linkages. For example, NCT numbers occasionally appear in records from other national trial registries; in such cases, we created edges of type \texttt{link\_to} to represent the equivalence between records referring to the same underlying trial across different sources. The statistics of all the relation types are in Supplementary Table~\ref{tab:stats_relation_table}.

\subsection{Disposition table}
Disposition reflects an essential aspect of a clinical trial: the design of experimental arms, including details such as the target treatment, administration schedule, duration, and dosage. To extract this information from ClinicalTrials.gov, we processed two complementary data sources: the trial design groups and their associated interventions. We began by loading and standardizing the relevant tab-separated files, namely \texttt{design\_groups}, \texttt{interventions}, and the linking table \texttt{design\_group\_interventions}. Key columns were normalized (e.g., renaming \texttt{id} to \texttt{design\_group\_id} or \texttt{intervention\_id}), and the data were merged using shared \texttt{nct\_id} and identifier fields. This enabled us to associate each design group with its corresponding intervention metadata. The resulting dataset provides, for each trial arm, a structured mapping that includes the intervention type (e.g., drug, device), group type (e.g., experimental, active comparator), treatment name, and descriptive text detailing the treatment protocol.

For publications, we extracted trial results in PICO format, where each unique pairing of the Population (P) and Intervention (I) elements is treated as a distinct disposition. From these pairs, we constructed corresponding trial arms that align with the disposition structure used in ClinicalTrials.gov. Each derived arm includes fields such as treatment name, description, and group type, allowing for consistent representation and integration across both registry-based and publication-based sources.

\subsection{Biomarkers table}

Biomarkers, particularly therapeutic biomarkers (ThMARs), are critical to the design and execution of clinical trials. ThMARs have become essential in both clinical development and practice across therapeutic areas. Five major types of ThMARs have been identified, each playing a distinct role in drug discovery and development: (1) \textit{Pharmacodynamic Biomarkers}, which ensure the pharmacological effect of a therapy; (2) \textit{Safety Biomarkers}, used to assess the likelihood or severity of therapy-induced toxicity; (3) \textit{Monitoring Biomarkers}, which guide ongoing clinical management by tracking patient status over time; (4) \textit{Predictive Biomarkers}, which help tailor therapies to individuals most likely to benefit; and (5) \textit{Surrogate Endpoints}, which serve as proxies to accelerate therapy approval.

We leveraged TheMarker~\cite{10.1093/nar/gkad862} as a curated reference source for therapeutic biomarkers (ThMARs) relevant to clinical trials. Our processing pipeline to identify and normalize biomarker mentions consisted of two stages. In the first stage, we employed a GPT-4o-mini model to extract candidate text spans from clinical trial descriptions, such as titles, abstracts, and outcome sections, that may contain therapeutic biomarkers. The model was prompted to directly extract potentially relevant mentions verbatim from the source text. For example, given a trial abstract, the model might identify spans such as ``HER2-positive breast cancer'' or ``PD-L1 expression level'' as candidate biomarkers.

In the second stage, we implemented a deterministic pattern matching module to normalize these candidate spans by linking them to standardized biomarker entries from TheMarker. This was achieved through a two-step matching strategy. First, we attempted an exact string match between the extracted phrase and the entries in a pre-indexed lookup table. If an exact match was found, the matched biomarker ID and corresponding metadata were recorded, and the phrase was classified with the match type \texttt{exact\_multi\_word}. If no exact match was found, we then attempted an order-invariant match by sorting the tokens in the candidate phrase and comparing them against a token-sorted index of biomarker terms. This allowed us to recover phrases with slight reordering of words, such as matching ``receptor estrogen alpha'' to ``estrogen receptor alpha,'' and was labeled as \texttt{exact\_multi\_word\_order\_invariant}.

Each successful match was associated with a structured biomarker representation that includes: (1) \texttt{biomarker\_name}, the canonical name of the matched biomarker; (2) \texttt{biomarker\_class}, describing the functional or molecular category of the biomarker (e.g., protein, receptor); (3) \texttt{biomarker\_type}, which corresponds to one of the five therapeutic biomarker types defined by TheMarker: \textit{Pharmacodynamic}, \textit{Safety}, \textit{Monitoring}, \textit{Predictive}, or \textit{Surrogate Endpoint}; and (4) \texttt{biomarker\_genes}, listing any specific genes associated with the biomarker when applicable. This structured mapping enables precise linkage between clinical trials and biomarker knowledge, facilitating downstream analyses on biomarker-driven trial design and therapeutic targeting.

Supplementary Figure~\ref{fig:biomarker_stats} shows the distribution of the most frequent biomarkers under each of the five therapeutic biomarker classes. In the Predictive (PRD) category, \textit{C reactive protein} and \textit{Tumor necrosis factor} appear most frequently, reflecting their widespread use in identifying responsive patient subpopulations. The {Surrogate (SUR)} category is heavily dominated by \textit{Blood pressure}, with over 60,000 mentions, followed by metrics such as \textit{Bone mineral density} and \textit{Serum creatinine}, commonly used as clinical endpoints in chronic disease trials. In the {Pharmacodynamic (PDY)} class, \textit{Low density lipoprotein} and \textit{Angiotensin converting enzyme} are among the most prominent, reflecting their roles in evaluating pharmacological activity, particularly in cardiovascular studies. For the {Safety (SAF)} biomarkers, \textit{Granulocyte macrophage colony-stimulating factor} and \textit{Hyaluronic acid} are the most frequently referenced, indicating their importance in monitoring adverse or toxic effects. Lastly, in the {Monitoring (MOI)} category, \textit{Epidermal growth factor}, \textit{HBV DNA}, and \textit{CEA (Carcinoembryonic antigen)} are among the top markers used for longitudinal assessment of disease progression or therapeutic response. Overall, this distribution highlights the heterogeneity in biomarker utility across different therapeutic contexts, with some biomarkers, such as \textit{C reactive protein} and \textit{Blood pressure}, being widely applicable across multiple trial types.

\subsection{Endpoints table}
Core Outcome Measurement Sets~\cite{dodd2018taxonomy,kirkham2022core} provide a standardized taxonomy for classifying clinical trial outcomes, comprising five high-level domains and their associated subdomains. We adopted this COMET taxonomy to categorize outcome measurements reported in clinical trials. To automate the classification process, we employed an in-context learning approach using a GPT-4o-mini model. For each trial, the model was prompted with structured instructions to first extract the \textit{primary outcome measurement} from the trial abstract or description, and then assign it to the appropriate COMET domain and subdomain. The full prompt used in this process is shown below.

\begin{lstlisting}
"""
You are a clinical research expert. Given a clinical trial abstract or description, extract the PRIMARY OUTCOME MEASUREMENT (usually only one or at most three), and classify them using the COMET taxonomy.

Follow these steps:
1. Extract the PRIMARY OUTCOME MEASUREMENT from the input text.
2. Classify the PRIMARY OUTCOME MEASUREMENT using the COMET taxonomy below.

COMET Taxonomy:

1. Mortality/survival  
   - No subdomains

2. Physiological/clinical  
   - Blood and lymphatic system outcomes  
   - Cardiovascular outcomes  
   - Endocrine outcomes  
   - Gastrointestinal outcomes  
   - General outcomes  
   - Immune system outcomes  
   - Infections and infestations outcomes  
   - Metabolic and nutritional outcomes  
   - Musculoskeletal and connective tissue outcomes  
   - Neoplasms and related outcomes  
   - Nervous system outcomes  
   - Pregnancy and childbirth outcomes  
   - Renal and urinary outcomes  
   - Reproductive system and breast outcomes  
   - Respiratory outcomes  
   - Skin and subcutaneous tissue outcomes  
   - Surgical and medical procedures outcomes  
   - Vascular outcomes

3. Life impact  
   - Delivery of care  
   - Global quality of life  
   - Perceived health status  
   - Personal circumstances  
   - Physical functioning  
   - Role functioning  
   - Social functioning  
   - Emotional functioning/wellbeing  
   - Cognitive functioning  
   - Sleep  
   - Fatigue

4. Resource use  
   - Hospital  
   - Need for further intervention  
   - Societal/carer burden  
   - Other resource use

5. Adverse events/effects  
   - Any subdomain

Return your output as a list of JSON objects, one per outcome, in the following format:

[
  {{
    "outcome": "<brief name of the primary outcome measurement>",
    "domain": "<COMET domain>",
    "subdomain": "<COMET subdomain or null>"
  }},
  ...
]

Input text:
{outcome_text}
"""
\end{lstlisting}

The prompt included the full taxonomy definitions as context, covering domains such as \textit{Mortality/Survival}, \textit{Physiological/Clinical}, \textit{Life Impact}, \textit{Resource Use}, and \textit{Adverse Events}, along with their respective subdomains. The model output was formatted as a list of JSON objects, each containing the outcome name, its mapped domain, and subdomain. This approach allowed for flexible, semantically aware interpretation of unstructured outcome descriptions, reducing the need for hand-crafted rules while achieving fine-grained alignment with the COMET classification. The result is a structured, interpretable dataset of standardized clinical trial endpoints that facilitates cross-trial comparisons, systematic reviews, and meta-analyses.

\subsection{Adverse events table}

The Adverse Event Table is systematically populated by processing the adverseEventsModule found within the resultsSection of structured JSON records from ClinicalTrials.gov. The primary aim of this table is to provide a comprehensive overview of the safety profiles reported in these clinical trials.

Adverse event information was extracted from ClinicalTrials.gov data using a custom rule-based parser. We process the JSON records to identify and structure reported adverse events. The key information extracted includes:

\begin{itemize}[leftmargin=*]
\item \textbf{Trial Arm Identification (I/C):} Distinct trial arms (or "event groups") are identified from the eventGroups list. The title and description of each group are combined to form a descriptive string for the intervention or comparator arm.
\item \textbf{Adverse Event Details:} For each arm, both "serious" (seriousEvents) and "other" (otherEvents) adverse event categories are processed. For each reported event, the parser extracts:
\begin{itemize}
\item The specific term used to describe the event (e.g., "Nausea").
\item The organSystem affected.
\item The frequency of the event within that specific arm, detailed as numAffected (number of participants experiencing the event) and numAtRisk (total participants in that arm). An event is only recorded for an arm if numAffected is greater than zero.
\end{itemize}
\item \textbf{Structured Data per Event Instance:} For each adverse event occurring in a specific trial arm, the following structured information is captured:
\begin{itemize}
\item The adverse event Term.
\item A comprehensive Description string that includes the organSystem, any associated notes, and the specific frequency of the event in that arm (formatted as "Frequency: X affected / Y at risk").
\item An is\_serious boolean flag to indicate if the event was classified as serious by the trial reporters.
\end{itemize}
\end{itemize}

The extracted and structured adverse event data is then assembled into a flattened tabular format. Each row in the final Adverse Event Table represents a single adverse event term as it was reported for a specific trial arm.

To enhance standardization and facilitate analysis, the reported adverse event Terms are mapped to the Medical Dictionary for Regulatory Activities (MedDRA). This mapping utilizes the MedDRA hierarchy files (LLT, PT, HLT, HLGT). The corresponding MedDRA code is then added to the table.

This systematically constructed Adverse Event Table, linked by NCTID and trial arm descriptions, serves as a valuable resource for safety analyses and complements the efficacy-focused Trial Result Table within the \datasetname~database.

\subsection{Data quality assessment}\label{appx:data_quality_assessment}
In the database building process, a series of automated tools are used. Including the ones based on pattern matching with medical ontologies, the ones based on medical named entity recognition (NER) and relation extraction (RE) tools, and the ones based on large language models. As a result, they inevitably have introduced some errors to the built databases. During the data quality assessment process, we identified two common errors.

The first type of error stems from ontology-based mapping, particularly when using tools like the RxNorm API. For instance, in trial \texttt{NCT06425341}, the drug intervention ``HRS-5635 Injection'' was incorrectly mapped to `HRS-1167'' due to name similarity. Such mismatches are common with pattern-matching-based medical ontologies and highlight the need for careful handling when using resources like UMLS or RxNorm.

Another common error arises from LLM-based extraction in complex fields like biomarkers. For example, in the clinical trial (PubMed ID: 17148749) titled ``Effects of PYY1-36 and PYY3-36 on appetite, energy intake, energy expenditure, glucose and fat metabolism in obese and lean subjects'', the model incorrectly identifies ``PYY'' as a biomarker (e.g., PYY mRNA). However, a closer reading shows that ``PYY'' refers to the intervention being tested, not a biomarker. Such cases highlight the difficulty of reliably extracting biomedical concepts using LLMs.

\clearpage

\section{Supplementary Figures}

\clearpage

\begin{figure}[t]
    \centering
    \includegraphics[width=0.8\linewidth]{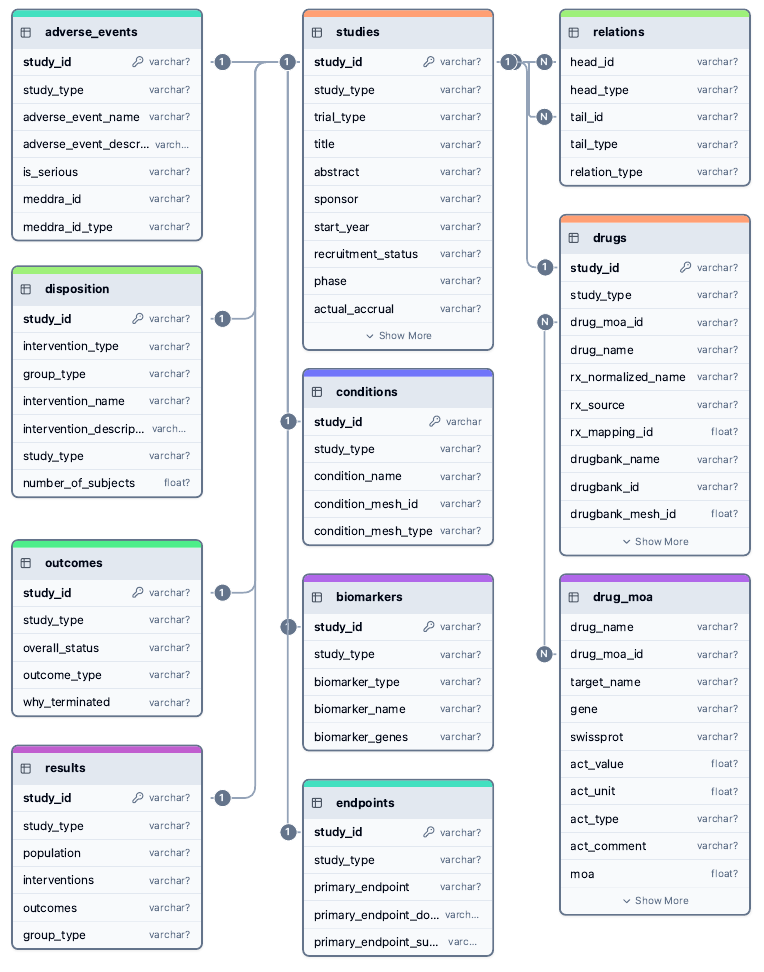}
    \caption{The schematic overview of the built database.}
    \label{fig:appx_database_schema}
\end{figure}

\clearpage

\begin{figure}[t]
    \centering
    \includegraphics[width=0.8\linewidth]{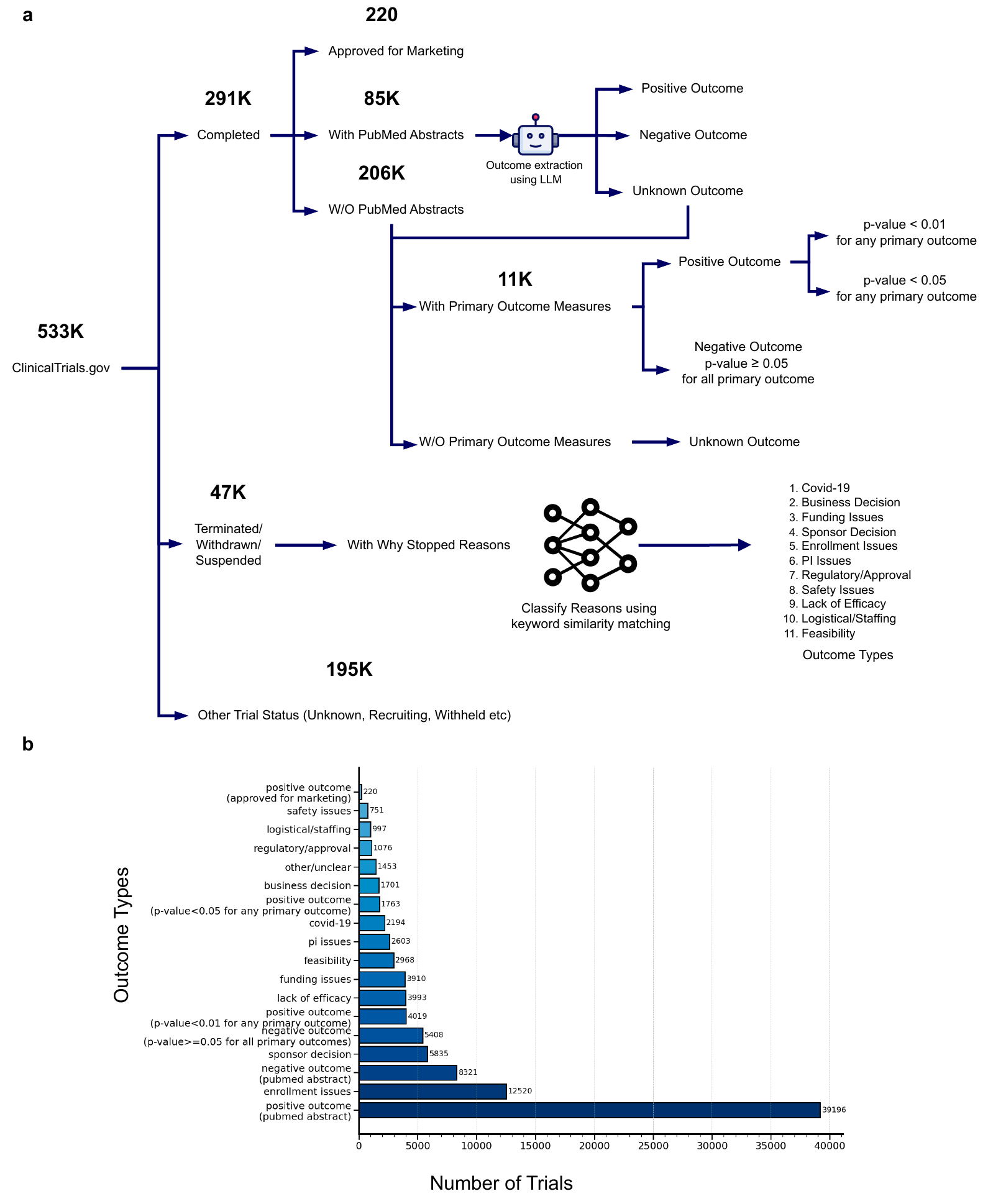}
    \caption{(a) Trial outcome extraction workflow. Completed trials are labeled using outcome measures and matched PubMed abstracts. Terminated trials are categorized by mapping stop reasons to predefined outcome types via keyword similarity. (b) Distribution of extracted outcome types on trials from ClinicalTrials.gov.}
    \label{fig:outcome}
\end{figure}

\clearpage

\begin{figure}
    \centering
    \includegraphics[width=0.8\linewidth]{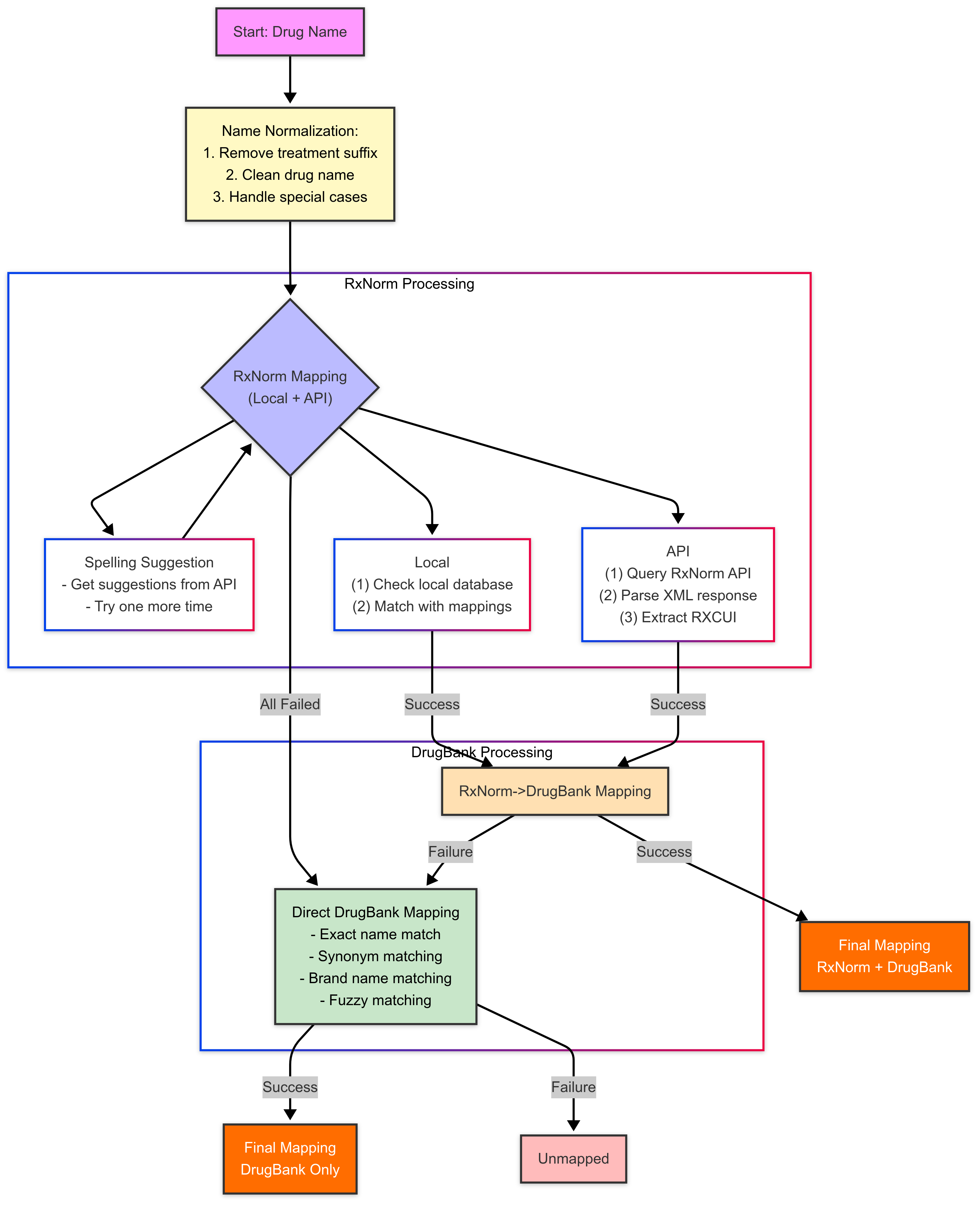}
    \caption{Pipeline for drug name normalization and standardization. Drug names are cleaned, mapped to RxNorm, and subsequently resolved to DrugBank identifiers using a combination of indirect and direct strategies.}
    \label{fig:drug_standard}
\end{figure}

\clearpage

\begin{figure}
    \centering
\includegraphics[width=\linewidth]{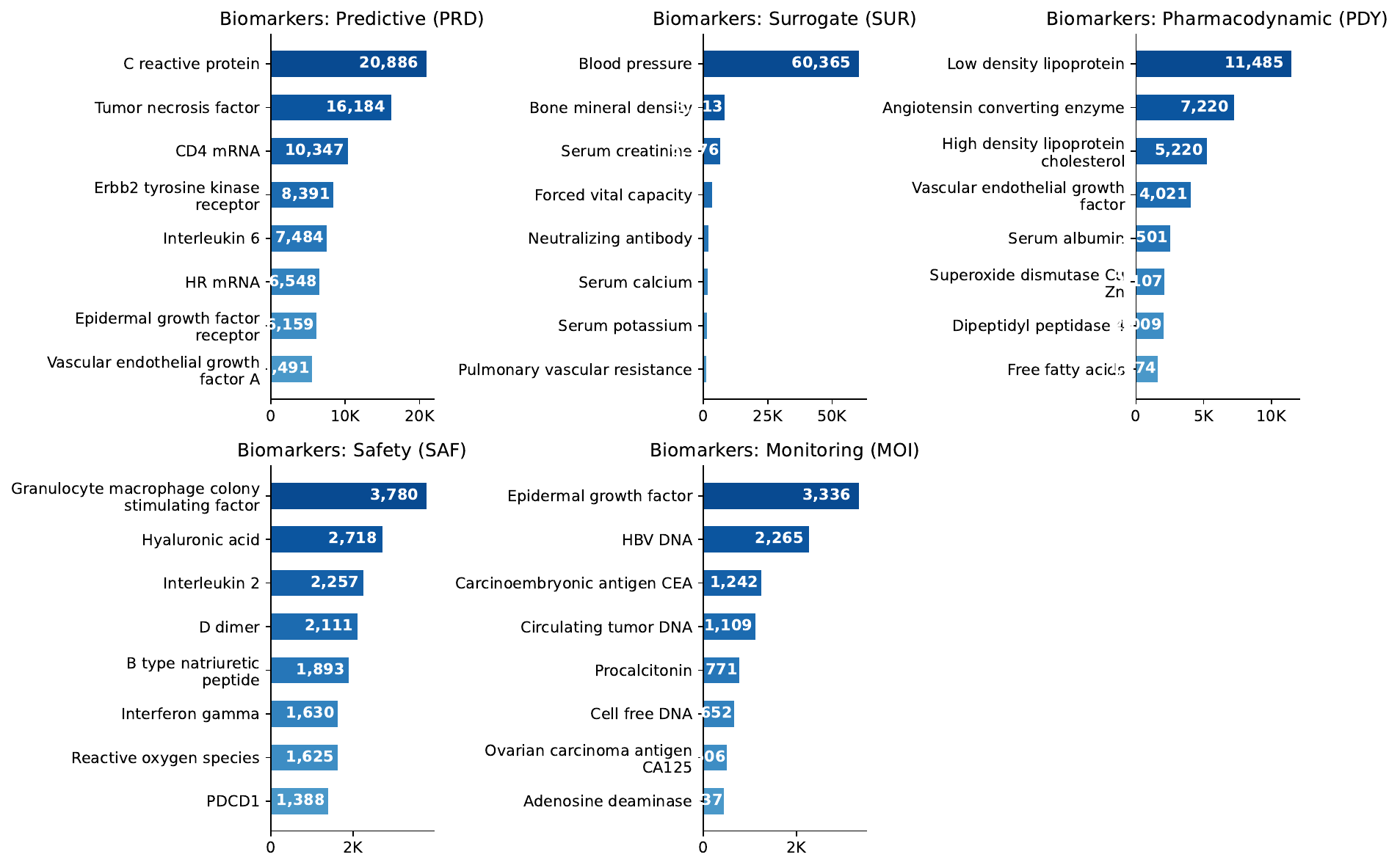}
    \caption{The distribution of the most frequent biomarkers under each of the five therapeutic biomarker classes.}
    \label{fig:biomarker_stats}
\end{figure}
\clearpage

\section{Supplementary Tables}

\clearpage

\begin{table}
\centering
\caption{List of data sources and the corresponding last crawl dates.}
\resizebox{\linewidth}{!}{
\begin{tabular}{lll}
\toprule
\textbf{Source} & \textbf{Data URL} & \textbf{Last Crawled} \\
\midrule
ANZCTR & \url{https://www.anzctr.org.au/} & April 1, 2025 \\
Brazilian Registry & \url{https://ensaiosclinicos.gov.br} & April 1, 2025 \\
ChiCTR  & \url{https://www.chictr.org.cn/searchprojEN.html} & April 1, 2025 \\
Dutch Registry & \url{https://www.onderzoekmetmensen.nl/en/} & April 1, 2025 \\
EU Clinical Trials & \url{https://www.clinicaltrialsregister.eu} & April 1, 2025 \\
DRKS  & \url{https://drks.de/search/en/results} & April 1, 2025 \\
Iranian Registry & \url{https://irct.behdasht.gov.ir/} & April 1, 2025 \\
ISRCTN  & \url{https://www.isrctn.com} & April 1, 2025 \\
Japan Registry & \url{https://center6.umin.ac.jp/cgi-open-bin/ctr_e/ctr_view.cgi} & April 1, 2025 \\
South Korea CRIS & \url{https://cris.nih.go.kr/cris/index/index.do} & April 8, 2025 \\
Pan African Registry & \url{https://pactr.samrc.ac.za/Search_v2.aspx} & April 8, 2025 \\
Sri Lanka CTR & \url{https://slctr.lk/trials} & April 8, 2025 \\
Thai Registry & \url{https://www.thaiclinicaltrials.org} & April 8, 2025 \\
REec  & \url{https://reec.aemps.es/reec/public/web.html} & April 9, 2025 \\
CTRI  & \url{https://ctri.nic.in/Clinicaltrials/advancesearchmain.php} & April 9, 2025 \\
PubMed & \url{https://ftp.ncbi.nlm.nih.gov/} & April 9, 2025 \\
\bottomrule
\end{tabular}
}
\label{tab:data_sources}
\end{table}

\clearpage

\begin{table}
\centering
\caption{Drug mapping statistics across data sources.}
\label{tab:drug_mapping_stats}
\resizebox{0.95\linewidth}{!}{
\begin{tabular}{lrrrrr}
\toprule
\textbf{Source} & \textbf{Total Drugs} & \textbf{RxNorm Mapped} & \textbf{RxNorm Rate (\%)} & \textbf{DrugBank Mapped} & \textbf{DrugBank Rate (\%)} \\
\midrule
PubMed & 953{,}361 & 579{,}035 & 60.74 & 241{,}879 & 25.37 \\
ClinicalTrials.gov & 366{,}214 & 290{,}371 & 79.29 & 273{,}509 & 74.69 \\
\bottomrule
\end{tabular}}
\end{table}

\clearpage

\begin{table}
  \centering
  \caption{Statistics of the relation types in the \texttt{relations} table.}
    \begin{tabular}{lll}
    \toprule
    Relation Type & Count & Percentage (\%) \\
    \midrule
    PubMed - cite - PubMed &            349,680  & 37.177 \\
    PubMed - exclude - PubMed &            237,076  & 25.205 \\
    PubMed - include - PubMed &            180,827  & 19.225 \\
    PubMed - linked\_to - ClinicalTrials.gov &            167,121  & 17.768 \\
    ClinicalTrials.gov - linked\_to - JAPAN &                  2,787  & 0.296 \\
    PubMed - include - ClinicalTrials.gov &                       954  & 0.101 \\
    ClinicalTrials.gov - linked\_to - KOREA &                       632  & 0.067 \\
    PubMed - cite - ClinicalTrials.gov &                       425  & 0.045 \\
    PubMed - exclude - ClinicalTrials.gov &                       406  & 0.043 \\
    ClinicalTrials.gov - linked\_to - GERMAN &                       331  & 0.035 \\
    ClinicalTrials.gov - linked\_to - ANZCTR &                       201  & 0.021 \\
    ClinicalTrials.gov - linked\_to - PAN\_AFRICAN &                          56  & 0.006 \\
    ClinicalTrials.gov - linked\_to - SRI\_LANKA &                          37  & 0.004 \\
    ClinicalTrials.gov - linked\_to - CHICTR &                          31  & 0.003 \\
    ClinicalTrials.gov - linked\_to - DUTCH &                          18  & 0.002 \\
    \bottomrule
    \end{tabular}%
  \label{tab:stats_relation_table}%
\end{table}%

\clearpage

\begin{table*}
    \centering
    \caption{Example instance of Study Search}
    \label{tab:study_search_example}    
    \begin{minipage}{0.95\columnwidth}
        \centering
        \begin{tcolorbox}[title=Source PubMed ID: 38078494]
            \textbf{Question:} Generate search query to find PubMed studies based on the systematic review setup as below.
\vspace{1em}

\begin{itemize}
    \item \textbf{Background}: Epilepsy is a common neurological disorder, affecting 0.5\% to 1\% of the population. In nearly 30\% of cases, epilepsy is resistant to currently available drugs. Pharmacological treatment remains the first choice to control epilepsy. Lamotrigine is a second-generation antiseizure medication. When used as an add-on (in combination with other antiseizure medications), lamotrigine can reduce seizures, but with some adverse effects.",
    \item \textbf{Objectives}: "To evaluate the benefits and harms of add-on lamotrigine, compared with add-on placebo or no add-on treatment in people with drug-resistant focal epilepsy.",
    \item \textbf{Selection criteria}: We included randomised controlled trials (RCTs) that investigated add-on lamotrigine versus add-on placebo or no add-on treatment in people of any age with drug-resistant focal epilepsy. We used data from the first period of eligible cross-over trials.
\end{itemize}
            \vspace{1ex}
            \textbf{Ground-truth studies (PubMed ID)}  "20696552", "2612495", "8937535", "10563619", "2498073", "2127016", "8232944", "8112232", "17938371", "18077797", "8505632", "8453943" 
        \end{tcolorbox}
    \end{minipage}
\end{table*}

\clearpage

\begin{table*}
    \centering
    \caption{Example instance of Study Screening}
    \label{tab:study_screening_example}
    \begin{minipage}{0.95\textwidth}
        \centering
        \begin{tcolorbox}[title=Source: PubMed ID: 38994714]
            \textbf{Background:} Cystic echinococcosis is a parasitic infection mainly impacting people living in low- and middle-income countries. Infection may lead to cyst development within organs, pain, non-specific symptoms or complications including abscesses and cyst rupture. Treatment can be difficult and varies by country. Treatments include oral medication, percutaneous techniques and surgery. One Cochrane review previously assessed the benefits and harms of percutaneous treatment compared with other treatments. However, evidence for oral medication, percutaneous techniques and surgery in specific cyst stages has not been systematically investigated and the optimal choice remains uncertain.

            \vspace{1ex}
            \textbf{Objective:} To assess the benefits and harms of medication, percutaneous and surgical interventions for treating uncomplicated hepatic cystic echinococcosis.

            \vspace{1ex}
            \textbf{Selection Criteria:} We included randomized controlled trials (RCTs) in people with a diagnosis of uncomplicated hepatic cystic echinococcosis of World Health Organization (WHO) cyst stage CE1, CE2, CE3a or CE3b comparing either oral medication (albendazole) to albendazole plus percutaneous interventions, or to surgery plus albendazole. Studies comparing praziquantel plus albendazole to albendazole alone prior to or following an invasive intervention (surgery or percutaneous treatment) were eligible for inclusion.

            \vspace{1ex}
            \textbf{Candidate Studies:}
            \begin{itemize}
                \item \textbf{PMID: 32382859} — Comparison of PAIR and catheterization techniques for CE1 and CE3a liver hydatid cysts.
                \item \textbf{PMID: 35708380} — [Title and abstract truncated for brevity.]
                \item \textbf{PMID: 16927868} — [Title and abstract truncated for brevity.]
                \item \textbf{PMID: 24464258} — [Title and abstract truncated for brevity.]
                \item \textbf{...} — (additional candidate studies omitted for space)
            \end{itemize}

            \vspace{1ex}
            \textbf{Screening Labels:}
            \begin{itemize}
                \item 32382859 — \textbf{Included}
                \item 35708380 — \textbf{Included}
                \item 16927868 — \textbf{Excluded}
                \item ...
            \end{itemize}
        \end{tcolorbox}
    \end{minipage}
\end{table*}

\clearpage

\begin{table*}
    \centering
    \caption{Example instance of Evidence Summarization}
    \label{tab:evidence_summarization}    
    \begin{minipage}{0.95\columnwidth}
        \centering
        \begin{tcolorbox}[title=Source PubMed ID: 10796093]
            \textbf{Question:} What were the findings regarding the effectiveness of bed rest in hospital for women with suspected impaired fetal growth compared to ambulatory management? Please answer this question based on the information provided below:

            \begin{center}
                {\{ title and abstract of included studies \}}
            \end{center}

            \vspace{1ex}
            \textbf{Options:}
            \begin{itemize}
                \item[A.] Bed rest in hospital significantly improved fetal growth parameters and neonatal outcomes.
                \item[B.] Bed rest in hospital significantly worsened fetal growth parameters and neonatal outcomes.
                \item[C.] There were no significant differences in fetal growth parameters and neonatal outcomes between bed rest in hospital and ambulatory management.
                \item[D.] Bed rest in hospital was found to be harmful to both the mother and the fetus.
            \end{itemize}

            \vspace{1ex}
            \textbf{Answer:} C
        \end{tcolorbox}
    \end{minipage}
\end{table*}

\clearpage

\begin{table*}
    \centering
    \caption{Sample Size Estimation Question}
    \label{tab:sample_size_question}
    \begin{minipage}{0.95\columnwidth}
        \centering
        \begin{tcolorbox}[title=Source NCTID: NCT06330298]
            \textbf{Question:} Here is the design of a clinical trial:

            Official Title: Improving Social Cognition and Social Behaviour in Various Brain Disorders.\\
            Included conditions: Stroke; Multiple Sclerosis; Brain Tumor.\\
            Study Arm groups: Experimental condition: Receives T-ScEmo Treatment (type EXPERIMENTAL; receives T-ScEmo between T0 and T1; interventionNames: [``Behavioral: Treatment social cognition and emotion regulation (T-ScEmo)'']); Waiting list group: Will be on waiting list instead of treatment (type NO INTERVENTION; duration T0–T1).\\
            Interventions: Type BEHAVIORAL; name Treatment social cognition and emotion regulation (T-ScEmo); description T-ScEmo is a multifaceted protocol to improve social cognition, regulation of social behaviour and participation (Westerhof et al., 2017, 2019); 15×1 h live sessions + 5 online practice sessions; involves significant other and homework; armGroupLabels: [Experimental condition: Receives T-ScEmo Treatment].\\
            Primary Outcomes: Change in social behaviour examined by proxy; difference T0→T2 on the Dysexecutive Questionnaire Social scales proxy version (DEX-Socproxy; Spikman et al., 2013; Westerhof-Evers, 2017); time frame: through completion (8–10 months).\\
            Please estimate the sample size based on: significance level ($\alpha=0.05$), power ($1-\beta=0.80$), and the group$\times$time interaction in a repeated measures analysis.

            \vspace{1ex}
            \textbf{Answer:} 84

            \vspace{1ex}
            \textbf{Explanation:} The study of Westerhof-Evers et al.\ [9] found that the DEX-Socproxy had an effect size of Cohen's d = 0.2 (Cohen's f = 0.10) for the comparison between T0 and T2. From these data, a small to medium effect size of Cohen's f = 0.15 was estimated for the current study. Therefore, the sample size is based on a power calculation for the main outcome measure (DEX-Socproxy) regarding the condition $\times$ time interaction (T0, T2) using repeated measures analysis with effect size Cohen's f = 0.15, $\alpha = 0.05$, and power = 0.80 [29], resulting in a sample size of $n = 62$. However, to ensure balanced distribution across patient groups, a total sample size of $n = 84$ will be used (42 patients per condition).
        \end{tcolorbox}
    \end{minipage}
\end{table*}

\clearpage

\begin{table*}[h]
    \centering
    \caption{Example instance of Trial Arm Selection}
    \label{tab:trial_arm_selection}    
    \begin{minipage}{0.95\columnwidth}
        \centering
        \begin{tcolorbox}[title=Source NCTID: NCT00000150]
            \textbf{Question:} Please select the arm or intervention descriptions that belong to the clinical trial below.

            \textit{Title:} Submacular Surgery Trials (SST)\\
            \textit{Brief Summary:} To determine whether surgical removal of subfoveal choroidal neovascularization (CNV) and associated hemorrhage in patients with age‐related macular degeneration (AMD), the ocular histoplasmosis syndrome (OHS), or idiopathic CNV stabilizes or improves vision more often than observation. To determine how surgical removal compared to observation of subfoveal CNV due to AMD, OHS, or idiopathic causes changes the patient’s perception of health‐ and vision‐related “quality of life,” as measured by telephone interview using the Medical Outcomes Survey Short Form-36 (MOS SF-36), the Hospital Anxiety and Depression Scale, and the National Eye Institute Visual Function Questionnaire (NEI VFQ-25). To determine whether randomized trials of surgery are warranted for patients with subfoveal CNV associated with AMD not suitable for laser treatment.

            \vspace{1ex}
            \textbf{Options:}
            \begin{itemize}
                \item[A.] Intervention Type: DRUG; Name: rhuFab V2 (ranibizumab); Assigned to Arm(s): N/A
                \item[B.] Arm Label: bevacizumab; Type: EXPERIMENTAL; Interventions: Drug: Bevacizumab \newline
                           Arm Label: control; Type: NO\_INTERVENTION; Interventions: N/A \newline
                           Intervention Type: DRUG; Name: Bevacizumab; Assigned to Arm(s): bevacizumab
                \item[C.] Intervention Type: PROCEDURE; Name: Subfoveal Choroidal Neovascularization Removal; Assigned to Arm(s): N/A
                \item[D.] Intervention Type: DRUG; Name: rhuFab V2 (ranibizumab); Assigned to Arm(s): N/A
            \end{itemize}

            \vspace{1ex}
            \textbf{Answer:} C
        \end{tcolorbox}
    \end{minipage}
\end{table*}

\clearpage

\begin{table*}
    \centering
    \caption{Example instance of Trial Eligibility Criteria Selection}
    \label{tab:trial_eligibility_selection}
    \begin{minipage}{0.95\columnwidth}
        \centering
        \begin{tcolorbox}[title=Source NCTID: NCT00004563]
            \textbf{Question:} Please select the eligibility criteria that correspond to the clinical trial below.

            \textit{Title:} Cyclophosphamide Versus Placebo in Scleroderma Lung Study\\
            \textit{Brief Summary:} To evaluate the efficacy and safety of cyclophosphamide versus placebo for the prevention and progression of symptomatic pulmonary disease in patients with systemic sclerosis.

            \vspace{1ex}
            \textbf{Options:}
            \begin{itemize}
                \item[A.] Inclusion: systemic scleroderma with active alveolitis on BAL fluid (neutrophilia >=3\% or eosinophilia >=2\%) on CT; symptom onset (excluding Raynaud's) within 7 years; FVC 45--85\% predicted; Mahler Dyspnea Index grade 2; Exclusion: DLCO <30\% predicted; smoking in past 6 months; pulmonary abnormalities; pulmonary hypertension requiring therapy; prednisone >10 mg/day or prior cyclophosphamide >4 weeks or >=2 IV doses; recent disease-modifying meds; ...
                \item[B.] Inclusion: idiopathic pulmonary fibrosis per ATS 2011, duration 6--48 months; age 40--80; \%FVC >=50\% <=90\%; \%DLCO >=30\% <=90\%; Exclusion: FEV1/FVC <0.8 post-bronchodilator; expected lung transplant within 1 year or on waiting list; known ILD cause; asthma/COPD history; active infection; ongoing IPF treatments; unstable cardiac/pulmonary disease in past 6 months; ...
                \item[C.] Inclusion: IPF diagnosis; age 40--80; FVC >=50\% predicted; DLCO >=35\% predicted; FVC or DLCO <=90\% predicted; no improvement in past year; walk 150 m in 6 min with sat >=83\% on <=6 L/min O2; Exclusion: unable PFT; obstructive lung disease or airway hyper-responsiveness; expected lung transplant within 72 weeks; active infection; liver disease; cancer or other condition likely death within 2 years; diabetes; pregnancy/lactation; substance abuse; long QT syndrome; other IPF treatment; inability to take study medication; withdrawal from IPF trials; ...
                \item[D.] Inclusion: IPF diagnosis; age 40--80; FVC >=50\% predicted; DLCO >=35\% predicted; FVC or DLCO <=90\% predicted; no improvement in past year; walk 150 m in 6 min with sat >=83\% on <=6 L/min O2; Exclusion: unable PFT; obstructive lung disease or airway hyper-responsiveness; expected lung transplant within 72 weeks; active infection; liver disease; cancer or other condition likely death within 2 years; diabetes; pregnancy/lactation; substance abuse; long QT syndrome; other IPF treatment; inability to take study medication; withdrawal from IPF trials; ...
            \end{itemize}

            \vspace{1ex}
            \textbf{Answer:} A
        \end{tcolorbox}
    \end{minipage}
\end{table*}

\clearpage

\begin{table*}
    \centering
    \caption{Example instance of Trial Outcome Measure Selection}
    \label{tab:trial_outcome_selection}
    \begin{minipage}{0.95\columnwidth}
        \centering
        \begin{tcolorbox}[title=Source NCTID: NCT00000170]
            \textbf{Question:} Please select the outcome measures that correspond to the clinical trial below.

            \textit{Title:} Amblyopia Treatment Study: Occlusion Versus Pharmacologic Therapy for Moderate Amblyopia\\
            \textit{Brief Summary:} To determine whether success rate with atropine vs. patching therapy in <7 y olds with strabismic or anisometropic amblyopia is equivalent; To develop precise estimates of treatment success rates; To identify factors associated with successful treatment; To collect detailed amblyopia course data; Extended follow-up: Primary—long-term visual acuity at age 10 and 15; Secondary—compare long-term acuity at age 10 and 15 between patching+care vs. atropine+care.

            \vspace{1ex}
            \textbf{Options:}
            \begin{itemize}
                \item[A.] Visual acuity of amblyopic eye by HOTV (<7 y) or ETDRS (>=7 y); measurement at 2 wk (strabismic/anisometropic/combined) and 4 wk (deprivation); ...
                \item[B.] Maximal improvement in visual acuity; ...
                \item[C.] Change in visual acuity from baseline to week 6 post-randomisation between arms; 6 wk; ...
                \item[D.] Visual acuity in amblyopic eye at 6 months; ...
            \end{itemize}

            \vspace{1ex}
            \textbf{Answer:} D
        \end{tcolorbox}
    \end{minipage}
\end{table*}

\clearpage

\begin{table*}
    \centering
    \caption{Example instance of Trial Completion Assessment}
    \label{tab:trial_completion_example}
    \begin{minipage}{0.95\textwidth}
        \centering
        \begin{tcolorbox}[title=Source NCTID: NCT02823470]
            \textbf{Study Title:} A Pilot Study to Evaluate the Use of Smart Adherence Technology to Measure Lumacaftor/Ivacaftor Adherence in CF Subjects Homozygous for the F508del CFTR Mutation

            \vspace{1ex}
            \textbf{Study Design:}
            \begin{itemize}
                \item Allocation: Randomized
                \item Intervention Model: Parallel
                \item Masking: None
                \item Primary Purpose: Treatment
                \item Phase: Phase 4
                \item Number of Arms: 2
            \end{itemize}

            \vspace{1ex}
            \textbf{Arms:}
            \begin{itemize}
                \item \textbf{Arm A} (Experimental): Activated smart device alerts and feedback\\
                \textit{LUM/IVA: LUM 400 mg q12h / IVA 250 mg q12h through Week 48.}
                \item \textbf{Arm B} (Experimental): De-activated smart device alerts/feedback features\\
                \textit{LUM/IVA: LUM 400 mg q12h / IVA 250 mg q12h through Week 48.}
            \end{itemize}

            \vspace{1ex}
            \textbf{Inclusion Criteria:}
            \begin{itemize}
                \item CF diagnosis with F508del-CFTR homozygosity, FEV1 $\geq$ 40\% predicted, consented, willing to comply.
                \item ... [truncated for brevity]
            \end{itemize}
            \textbf{Exclusion Criteria:}
            \begin{itemize}
                \item Hepatic impairment, mechanical ventilation, pregnancy, recent investigational drug use, prior/current use of commercial LUM/IVA.
                \item ... [truncated for brevity]
            \end{itemize}

            \vspace{1ex}
            \textbf{Outcome:} \texttt{Terminated}

            \textbf{Reason for Termination:} Enrollment issues — study terminated early due to slow recruitment of patients.
        \end{tcolorbox}
    \end{minipage}
\end{table*}


\end{document}